\pdfoutput=1

\documentclass[11pt]{article}

\usepackage[]{acl}

\usepackage{times}
\usepackage{latexsym}

\usepackage[T1]{fontenc}

\usepackage[utf8]{inputenc}

\usepackage{microtype}

\usepackage{inconsolata}
\usepackage{amssymb,amsmath}
\usepackage{booktabs}
\usepackage{multirow}
\usepackage{graphicx}
\usepackage{soul}
\usepackage{todonotes}
\usepackage{arydshln}
\usepackage{hyperref}

\usepackage{svg}
\newcommand{\task}{\texttt{MemeMQA}}
\newcommand{\dataset}{\texttt{MemeMQACorpus}}
\newcommand{\exhvv}{\texttt{ExHVV}}
\newcommand{\model}{\texttt{ARSENAL}}
\usepackage{enumitem}
\newlist{tabenum}{enumerate}{1}

\title{\task: Multimodal Question Answering for Memes \\via Rationale-Based Inferencing}

\author{Siddhant Agarwal$^1$\thanks{\;\;denotes equal contribution}, Shivam Sharma$^{2,3}$\footnotemark[1], \textbf{Preslav Nakov$^4$, Tanmoy Chakraborty$^2$}\\
  $^1$Indraprastha Institute of Information Technology Delhi, India\\
  $^2$Indian Institute of Technology Delhi, India $^3$Wipro R\&D (Lab45), India \\  
  $^4$Mohamed bin Zayed University of Artificial Intelligence, UAE \\  
  \small\texttt{siddhant20247@iiitd.ac.in}, \small\texttt{\{shivam.sharma, tanchak\}@ee.iitd.ac.in}, \small\texttt{preslav.nakov@mbzuai.ac.ae}}

\begin{document}
\maketitle
\begin{abstract}
Memes have evolved as a prevalent medium for diverse communication, ranging from humour to propaganda. With the rising popularity of image-focused content, there is a growing need to explore its potential harm from different aspects. Previous studies have analyzed memes in closed settings -- detecting harm, applying semantic labels, and offering natural language explanations. To extend this research, we introduce \task, a multimodal question-answering framework aiming to solicit accurate responses to structured questions while providing coherent explanations. We curate \dataset, a new dataset featuring $1,880$ questions related to $1,122$ memes with corresponding answer-explanation pairs. We further propose \model, a novel two-stage multimodal framework that leverages the reasoning capabilities of LLMs to address \task. We benchmark \task\ using competitive baselines and demonstrate its superiority --  $\sim$$18\%$ enhanced answer prediction accuracy and distinct text generation lead across various metrics measuring lexical and semantic alignment over the best baseline. We analyze \model's robustness through diversification of question-set, confounder-based evaluation regarding \task's generalizability, and modality-specific assessment, enhancing our understanding of meme interpretation in the multimodal communication landscape.\footnote{\textcolor{red}{CAUTION: Potentially sensitive content included; viewer discretion is requested.}}
\end{abstract}

\section{Introduction}
Memes offer an accessible format for impactful information dissemination for everyone without conventional dependencies of proper formatting or formal language. 
It provides an easy opportunity for novice content creators and seasoned professionals to propagate information that may sometimes be harmful to the general audience, especially in the age of Internet virality. Previous work has explored aspects such as harmfulness in various forms, such as hate speech \cite{kiela_hateful_2020}, cyber-bullying \cite{Survey:2022:Harmful:Memes}, and offensive languages \cite{shang2021aomd}, of memes, typically in a black-box setting. 

\begin{figure}[t!]
    \centering
    \includegraphics[width=\columnwidth]{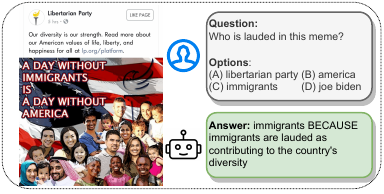}
    \caption{The \task\ task: Given an input meme and multiple choices, identify the correct answer and justify.}
    \label{fig:intro figure}
    \vspace{-5mm}
\end{figure}

Memes, with their appealing format and influential nature on social media, necessitate the modeling of complex aspects like harmfulness, targeted social groups, and offensive cues to assess their narrative framing and ensure online content safety. Their growing prevalence as a key medium for information dissemination poses significant societal challenges. A question-answering setup, particularly open-ended or instruction/response formats, offers a user-friendly method for probing models about the potential harmfulness of memes and understanding their responses. This approach enhances model interpretability and serves as an effective tool for content moderation.

In this work, we explore contextualized semantic analysis of memes by introducing a novel multimodal task, called \task\ (c.f. Fig.~\ref{fig:intro figure} for demonstrative example), which is formulated as follows: Given a meme and a structured question about the semantic role assigned to various entities, (a) deduce the correct answer entity from a set of multiple options, while also, (b) generating succinct explanations towards the answer choice.

Building on the work of \cite{sharma-etal-2022-findings}, we explore the narrative framing of entities like well-known individuals and political figures in online memes. This research is especially important during critical events like elections or pandemics, where the risk of spreading harmful content such as hate speech and misinformation increases, highlighting the need for effective moderation. We adopt terms like `hero', `villain', and `victim' from \cite{sharma-etal-2022-findings} to analyze memes' intentions of victimization, glorification, and vilification. Our goal is to deepen the understanding of these memes and contribute to making social media safer. The MemeMQA framework is designed to assist social media users and fact-checkers in evaluating the harmfulness of memes, enabling them to ask questions and receive accurate, informed responses.


Analyzing memes in \task\ is complex due to their nuanced meanings that demand advanced reasoning, including common sense, world knowledge, and cultural understanding. For instance, the meme in Fig.~\ref{fig:intro figure} could simultaneously highlight the role of immigrants in America and promote the Libertarian Party. To correctly answer the question ``Who is lauded in this meme?'', it's essential to grasp the meme's key themes and the implied message about immigrants enriching diversity, which directly glorifies them. Therefore, ``immigrants'' is the most suitable answer in this context, rather than ``Libertarian Party'' or ``America'', which, despite being referenced positively, would lead to an incorrect conclusion.


In summary, we introduce a new task for answering and explaining multiple-choice questions about political memes, creating a dataset (\dataset) with $1,880$ questions for $1,122$ memes using \exhvv\ dataset \cite{Sharma_Agarwal_Suresh_Nakov_Akhtar_Chakraborty_2023}. We benchmark \dataset\ with various unimodal and multimodal baselines, including recent multimodal LLMs, and propose \model, a novel modular approach that leverages multimodal LLM reasoning capabilities. \model~includes rationale, answer prediction, and explanation generation modules. We analyze and compare the performance of \model\ against these baselines, highlighting its strengths and limitations. Our contributions are summarised as follows\footnote{Supplementary accompanies the source codes and sample dataset.}:
\begin{enumerate}[leftmargin=*,noitemsep]
    \item \textbf{\task}: A novel task formulation that introduces a multimodal question-answering setup in the context of memes.
    \item \textbf{\dataset}: An extension of a previously available dataset to introduce a set of diverse questions and multiple choice settings for \task.
    \item \textbf{\model}: A multimodal modular framework system architecture that leverages multimodal LLM generated rationales for \task.
    \item An exhaustive study in the form of benchmarking, prompt evaluations, detailed analyses of diversified questions, confounding-based cross-examination, implications of multimodality and limitations of the proposed solution.
\end{enumerate}

\section{Related Work}
This section provides a concise coverage of prominent studies on meme analysis, while also reviewing contemporary works within the domain of Visual Question Answering. Finally we consolidate our assessment of the current state-of-the-art in Multimodal LLMs. 

\paragraph{Studies on Memes.}
Recent collaborative efforts encompass diverse meme analysis aspects, including entity identification \cite{sharma-etal-2022-findings, Prakash_2023}, emotion prediction \cite{sharma-etal-2020-semeval} and notably, hateful meme detection \cite{kiela_hateful_2020,zhou2021multimodal} through methods like fine-tuning Visual BERT, UNITER \cite{li2019visualbert,chen2020uniter}, and dual-stream encoders \cite{muennighoff2020vilio,sandulescu2020detecting,lu2019vilbert,zhou2020unified,tan2019lxmert}. Further studies address anti-semitism, propaganda, harmfulness \cite{chandra2021subverting,dimitrov-etal-2021-detecting,pramanick-etal-2021-momenta-multimodal,suryawanshi-chakravarthi-2021-findings,Prakash_2023,sharma-etal-2022-disarm}, while recent research explores multimodal evidence prediction, role-label explanations \cite{Sharma_Agarwal_Suresh_Nakov_Akhtar_Chakraborty_2023}, and semantic analysis of hateful memes \cite{hee2023decoding, cao-etal-2022-prompting, ukraine_memes}. Most of these studies are constrained by the schema and quality of the annotations while limiting the open-ended probing of memetic phenomena.

\paragraph{Visual Question Answering (VQA).}
This subsection explores the evolution of VQA research. Initial pioneering work by \citet{VQA} emphasized open-ended questions and candidate answers. Subsequent studies introduced variations, including joint image and question representation, to classify answers \cite{VQA}. Researchers further explored cross-modal interactions using various attention mechanisms, such as co-attention, soft-attention, and hard-attention \cite{VQA_Parikh1, anderson2018bottom, eccvattention2018}. Notably, efforts were made to incorporate common-sense reasoning \cite{zellers2019recognition, wu2016ask, wu2017image, marino2019ok}. Models like UpDn \cite{anderson2018bottom} and LXMERT \cite{tan2019lxmert} harnessed non-linear transformations and Transformers for VQA, while addressing language priors \cite{clark-etal-2019-dont,zhu2020prior}. In a standard Visual-Question-Answering framework, an image is presented alongside a related question and, depending on the setup, multiple-choice options. Memes, however, introduce a more complex layer, combining images with frequently mismatched textual content, making the task more challenging and far from straightforward.


\paragraph{Multimodal Large Language Models.}
The rise of large language models (LLMs) like ChatGPT \cite{chatgpt2022}, GPT4 \cite{openai2023gpt4}, Bard \cite{bard2023}, LLaMA \cite{touvron2023llama}, Vicuna \cite{vicuna2023}, etc., has brought significant advancements in natural language understanding and reasoning. Their affinity towards multimodal augmentation is also reflected for visual-linguistic grounded tasks. Such models augment LLMs via fusion-based \textit{adapter} layers, to excel at various tasks, from VQA to multimodal conversations \cite{alayrac2022flamingo,awadalla2023openflamingo,liu2023visual,openai2023gpt4,zhu2023minigpt4,gong2023multimodalgpt,zhao2023chatbridge}. However, existing multimodal LLMs like LLaVA \cite{liu2023visual}, miniGPT4 \cite{zhu2023minigpt4}, and multimodalGPT \cite{gong2023multimodalgpt} exhibit limitations in grasping nuances like \textit{sarcasm} and \textit{irony} in visual-linguistic incongruity seen in memes. Although few similar works address meme-related tasks, it's mainly limited to visual-linguistically grounded settings of caption generation and VQA \cite{hwang2023memecap}. For a more comprehensive range of tasks, they exhibit limitations inherent to LLMs, like \textit{pre-training biases} and \textit{hallucinations} \cite{zhao2023chatbridge}. 

The dual objectives of \task, encompassing answer prediction and explanation generation, present unique challenges. Existing methods fall short, including the Multimodal CoT (MM-CoT) model \cite{zhang2023multimodal}, a two-stage framework combining DETR-based visual encoding \cite{carion2020endtoend} and textual encoding/decoding from \texttt{unifiedqa-t5-base}\footnote{\url{https://huggingface.co/allenai/unifiedqa-t5-base}}. MM-CoT excels in answer prediction but falters in explanations. Instruction-tuned multimodal LLMs like LLaVA, InstructBLIP \cite{dai2023instructblip}, and miniGPT4 show promise in understanding meme semantics but struggle with question-specific accuracy, prioritizing broader meme context over precise answers. In this work, our focus is on addressing challenges pertaining to complex visual-semantic reasoning, posed by \task\ task while considering limitations in current multimodal LLMs and neural reasoning setups for question-answering.

\section{The \dataset\ Dataset}
Current meme datasets typically encompass either categorical labels \cite{kiela2020hateful,pramanick-etal-2021-detecting,shang2021aomd} or their associated explanations \cite{Sharma_Agarwal_Suresh_Nakov_Akhtar_Chakraborty_2023}. Although conventional Visual Question Answering (VQA) \cite{VQA,VQA_Parikh1} frameworks exist, they lack the nuanced complexity of memes. These include tasks like detection, segmentation, conditional multimodal modeling (such as caption generation, visual question answering, and multiple-choice VQA), and strong visual-linguistic integration (e.g., setups similar to MS COCO for question-answering that focus on common-sense and objective reasoning) \cite{VQA,VQA_Parikh1}. While these areas present their distinct challenges and mark a significant line of inquiry within the intersecting realms of computer vision and natural language processing (multimodality), they fall short of addressing the complexities of multimodal \textit{reasoning, abstract idea representation, and the nuanced use of language mechanisms like puns, humor, and figures of speech, etc.} These elements are often integral to memes. This oversight has generally curtailed the effectiveness of existing multimodal approaches \cite{pramanick-etal-2021-momenta-multimodal} in capturing the \textit{nuanced complexities} inherent to memes.

To address this gap, we introduce \dataset, a novel dataset designed to \textit{emulate} free-form questioning and multiple-choice answering. Given the overwhelming diversity of possible question-answer pairs for the multifarious phenomena presented in memes, we supplement \exhvv\ \cite{Sharma_Agarwal_Suresh_Nakov_Akhtar_Chakraborty_2023}, an existing multimodal dataset consisting of natural language explanations for connotative roles for \textit{three} entity types - \textit{heroes, villains}, and \textit{victims}, across $4,680$ instances for $3K$ memes, with automatically constructed, structured questions. This expansion aims to emulate the intricacies of meme interpretation and communication via a \textit{QnA} setup.

\begin{figure}[!t]
    \centering
    \includegraphics[width=\columnwidth]{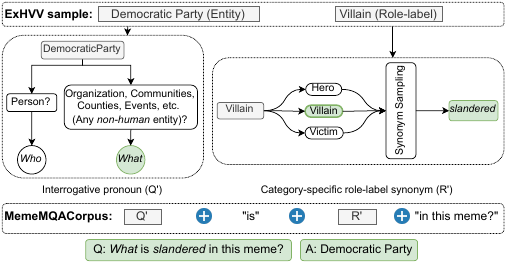}
    \caption{A schematic diagram showing question-answer construction process in \dataset, using entity and role-label information from \exhvv.}
    \label{fig:dataproc}
\end{figure}


Our approach entails crafting structured question sets for distinct role categories -- hero, villain, and victim, each linked to unique entities featured in \exhvv's memes, as depicted in Fig. \ref{fig:dataproc}. The central goal is to create role-based queries that precisely elicit only one entity as the correct answer within a multiple-choice setting. For instance, when an entity such as \emph{Democratic Party} is labeled as a \emph{villain} in a meme, along with the availability of a corresponding explanation from the \exhvv\ dataset, we formulate the question: ``\emph{What is slandered in this meme?}'' (c.f. Fig. \ref{fig:dataproc}). With \emph{Democratic Party} as the correct choice, distractive choices for answer options are selected based on entities referred within the meme, sampled randomly from the ones not sharing the role label with the ground truth. Sampling occurs from the entire training corpus in cases with insufficient valid entity choices. 
\begin{figure}[t!]
    \centering
    \includegraphics[width=\columnwidth]{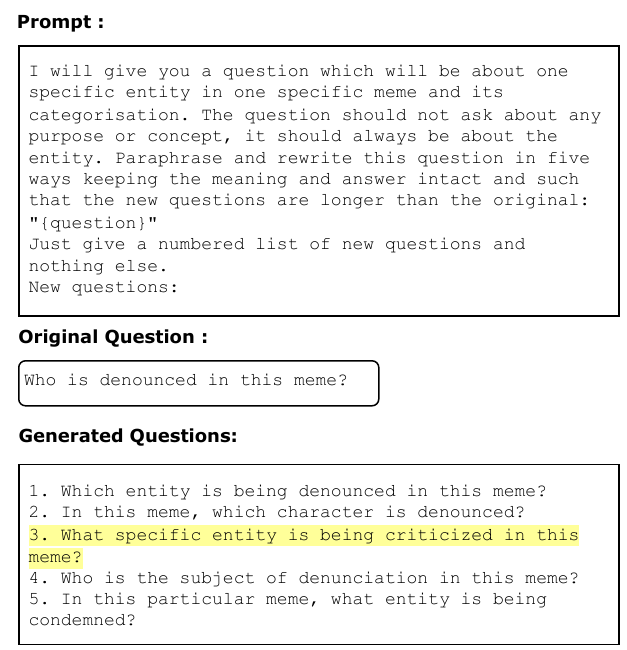}
    \caption{Description of the prompting setup for free-form synthetic question generation using the LLM, \texttt{Llama-2-7b-chat}. The randomly chosen question option is highlighted in yellow.}
    \label{fig:prompting_llama2}
\end{figure}
Additionally, to emulate free-form questioning by increasing the question variability, we incorporate various synonyms of role labels ({hero}, 
 {villain}, and {victim}). The role-synonym mapping and their proportional breakdown, integral to constructing queries in \dataset, are shown in Table~\ref{tab:synonym}. Our curation effort encompasses $1,880$ meme-question pairs, corresponding to $1,122$ distinct memes about \textit{US Politics}. This domain choice is based on diversity in the entity distribution across different roles compared to the other subset (on \textit{Covid-19}) of \exhvv\ dataset. To further examine the robustness of different modeling approaches, we curate additional variants of \dataset, with (a) Question Diversification, and (b) Confounding Analysis, the details of which are presented in Sec. \ref{sec:rob}.
 As our question enhancement approach is automated, we achieve valid questions seamlessly linked to \exhvv\ instances, relying on the pre-existing annotations.

\subsection{Prompting for Question Diversification}
\label{app:sec:questdiv}

To achieve diversity in the framing of the original questions, a pre-trained LLM, \texttt{Llama-2-7b-chat}, is utilised for inferencing via zero-shot prompting. In this setting, the LLM is provided a context about the setting of the question which is followed by asking the model to rewrite the question in multiple ways without changing the meaning of the quetion. This ensures that the original meaning and, hence, the validity of the original option set remains intact. One out of the five rephrased questions provided by the LLM is then chosen at random. This chosen question replaces the original question in \dataset, extending the questioning style of \dataset\ to emulate free-form question answering more closely.

\begin{figure}[t!]
    \centering
    \setlength{\fboxsep}{1pt}%
    \includegraphics[width=\columnwidth]{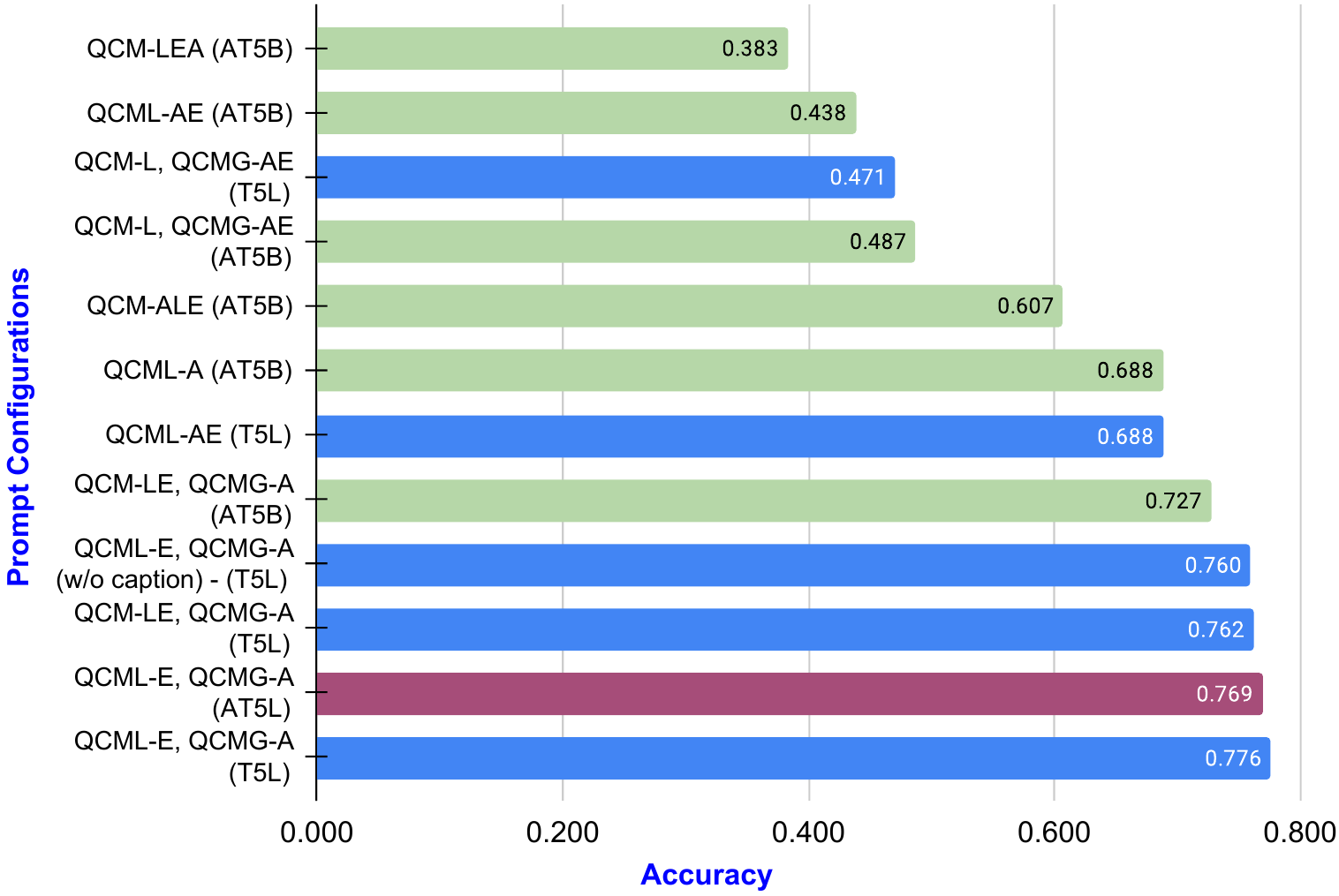}
    \caption{Comparison of various prompt configurations examined. Bar color scheme -- Green: \texttt{unifiedqa-t5-base}, Magenta: \texttt{unifiedqa-t5-large}, and Blue: \texttt{t5-large}.}
    \label{fig:prompt}
\end{figure}

\begin{table}[t!]
\centering
\resizebox{\columnwidth}{!}{%
\begin{tabular}{@{}ccp{6cm}}
\toprule
\textbf{Role-label} & \textbf{Counts (\%)} & \multicolumn{1}{c}{\textbf{Synonyms}} \\ \midrule
hero & 222 (17\%) & glorified, praised, lauded, idealized \\\hdashline
\multirow{2}{*}{villain} & \multirow{2}{*}{1297 (59\%)} & vilified, berated, slandered, defamed, denounced, disparaged, maligned \\\hdashline
\multirow{2}{*}{victim} & \multirow{2}{*}{361 (24\%)} & victimised, exploited, taken advantage of, scapegoated \\ \bottomrule
\end{tabular}%
}
\caption{The synonyms used, corresponding to the role-labels \textit{hero}, \textit{villain}, and \textit{victim} (and their proportions) as part of the \dataset\ dataset.}
\label{tab:synonym}
\end{table}

\begin{figure*}
    \centering
    \includegraphics[width=\textwidth]{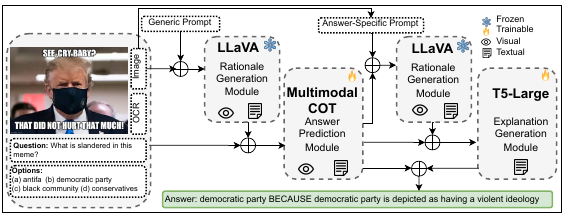}
    \caption{A schematic diagram of \model\ for the \task\ task ($\bigoplus$: fusing the information via concatenation).}
    \label{fig:enter-label}
\end{figure*}

\section{The \model\ Model}
\label{sec:proposed_approach}
Prior to exploring an effective design towards addressing \task, we analyze different prompting configurations using meme-based inputs to determine the optimal strategy. This section begins by outlining the optimal prompting strategy, then details the structural aspects of \model.


\paragraph{Prompting Configurations:}

In multimodal question-answering with CoT reasoning \cite{zhang2023multimodal}, the setup includes a question, context (text associated with an image), options, lecture (detailed generic context), explanation (a concise contextual statement), answer, and intermediate generated text.\footnote{Unless stated otherwise, these are typically abbreviated as \texttt{QCMLEAG} -- question \texttt{Q}, context \texttt{C}, multiple options \texttt{M}, lecture \texttt{L}, explanation \texttt{E}, answer \texttt{A}, and generated intermediate text \texttt{G}.} Prompt configurations are represented as \texttt{input}$\rightarrow$\texttt{output}, combining elements from \texttt{QCMLEAG}. Prior one-stage approaches (\texttt{QCM$\rightarrow$LA} or \texttt{QCM$\rightarrow$AL}) have limitations, prompting a two-stage setup with improved performance \cite{zhang2023multimodal}. Since \task\ involves more complex reasoning than ScienceQA \cite{lu2022learn}, we first examine $11$ prompt configurations for \task, with lectures (\texttt{L}) as detailed role definitions, using one/two-stage methods and base models \texttt{unifiedqa-t5-base/large} (AT5B/L) and \texttt{t5-large} (T5L). Our findings (c.f. Fig. \ref{fig:prompt}) corroborate the applicability of the two-stage framework for \task.\footnote{Refer App. \ref{app:prompt} for more details on \textit{Prompting Configuration Assessment}.}

\subsection{System Architecture}
Our input consists of three parts: (i) the meme image, $Meme_{I}$, (ii) the OCR Text, \begin{math}Meme_{T}\end{math}, and (iii) the question \texttt{Q} with its corresponding multiple options \texttt{M}. The expected output consists of two parts -- (i) the answer, \begin{math}Y_{answer}\end{math}, and (ii) the explanation, \begin{math}Y_{exp}\end{math}.
We propose a multi-stage setup for \model~to leverage individual strengths of MM-CoT and multimodal LLMs towards the overall objective of \task. The framework is a two-stage process consisting of answer prediction and explanation generation. It has a modular design, incorporating LLM-inferred rationale in both stages. The initial stage includes two steps: generating an intermediate rationale and predicting the answer, while the second stage focuses on generating explanations.

\paragraph{Rationale Generation:}
We curate ``generic rationale,'' $R_{generic}$, offline in the \textit{first} stage to provide semantic information about the meme in a textual form, which is generally not captured well by the OCR information alone. $R_{generic}$ is developed using the multimodal LLM, LLaVA-7B  using zero-shot inference with the prompt, $P_{generic}$ \textit{``Explain this meme in detail.''} The multimodal {LLaVA-7B} LLM is built on the base LLM, \texttt{Vicuna-7B}. The $R_{generic}$ thus generated captures relevant semantic information deemed useful for providing semantic clues in further stages of the proposed framework. This process can be expressed as follows:
\begin{equation}
\small{
    R_{generic} = Model_{LLaVA}(Meme_{I}, P_{generic})}
\end{equation}

In the \textit{second} stage, {LLaVA-7B} model is again used to generate an ``answer-specific rationale,'' $R_{specific}$, by prompting the model with a combination of the answer generated in the first stage and an answer-specific prompt, $P_{specific}$. $P_{specific}$ is of the form -- \texttt{``How is [answer] [rephrased question]''}, where the rephrased question is framed by removing the first two words of the question. For example, for a question, \texttt{Q}, given as \textit{``Who is victimised in this meme?''} with the answer \textit{`Joe Biden'}, the rephrased question would be given as \textit{``How is Joe Biden victimised in this meme?''}. This is represented as,
\begin{equation}
    \small{R_{specific} = Model_{LLaVA}(Meme_{I}, P_{specific})}
\end{equation}
\paragraph{Stage 1 - Answer Prediction:}
This stage
implements 
  {Mutimodal CoT} model with two-stage training. It uses the T5-large model with the prompting strategy of \texttt{QCM$\rightarrow$LE} followed by \texttt{QCMG$\rightarrow$A}. The model is provided with visual data in the form of embeddings obtained from the  {DETR} model. These embeddings are used by adding a gated-cross attention layer in the encoder stack of the T5 model as follows: \begin{math}
    H_{fuse} = (1 - \lambda) \cdot H_{language} + \lambda \cdot H^{attn}_{vision}
\end{math}
, where $\lambda$ is the sigmoid-activated output of fused image+text embeddings, $H_{language}$: text-input embeddings and $H^{attn}_{vision}$: output of text+vision cross-attention. We provide this model with additional contextual cues from $R_{generic}$. The model is then fine-tuned on \dataset, with the first step of training for five epochs being a text generation task with the objective of generating text, \texttt{G}, of the form \begin{math}R_{generic} \;[SEP]\; Y_{exp}\end{math}.

\begin{equation}
    \small{G = Model_{mm-cot}(Meme_{I}, Meme_{T}, Q, M)}
\end{equation}
 This is followed by another training step for five epochs to fine-tune the model for generating  $Y_{answer}$.

\begin{equation}
    \small{Y_{answer} = Model_{mm-cot}(Meme_{I}, Meme_{T}, Q, M, G)}
\end{equation}     

\paragraph{Stage 2 - Explanation Generation:}
The second stage focuses on generating an explanation for the answer obtained from the previous stage. To this end, the {LLaVA-7B} model is used again for its superior reasoning capacity to generate an \textit{answer-specific} rationale, $R_{specific}$. 
This provides us with a highly informative rationale that focuses specifically on the chosen answer and provides a highly relevant explanation. However, this generation lacks the structure and the conciseness of the expected explanation. To this end, $R_{specific}$ is provided along with the question and correct answer to a unimodal {T5-large} model for text-to-text generation. This {T5-large} model is fine-tuned for two epochs in a text-to-text generation setting for generating the expected explanation. The prompt \begin{math}P_{T5}\end{math}, given to the T5 model, is \texttt{``Summarize the explanation for question based on the answer''}. The task of T5 is as follows:
\begin{equation}
    \small{Y_{exp} = Model_{T5}(P_{T5}, Q, Y_{answer}, R_{specific})}\label{eq:5}
\end{equation}
While we fine-tune it for the conditional generation objective and obtain the T5-decoder's language modeling loss $\mathcal{L}^{\text{EXP}} = -\log(p_{y_{t}}) = -\log(p(y_{t}|y_{< t}))$. 
The resultant explanation is combined with the previously obtained answer to obtain our final result of the form -- \texttt{``Answer: [answer] BECAUSE [explanation]''}.

\begin{table*}[t!]
\centering
\resizebox{\textwidth}{!}{%
\begin{tabular}{@{}clccccccc@{}}
\toprule
\textbf{Type} & \multicolumn{1}{c}{\textbf{Models}} & \textbf{Accuracy} & \textbf{BLEU-1} & \textbf{BLEU-4} & \textbf{ROUGE-L} & \textbf{METEOR} & \textbf{CHRF} & \textbf{BERTScore} \\ \cmidrule(r){1-3} \cmidrule(l){4-9}
\multirow{4}{*}{UM} & UM.TEXT.T5 & 0.53 & \ul{0.59} & \ul{0.15} & 0.44 & 0.41 & 0.35 & 0.901 \\
& UM.TEXT.GPT3.5 & 0.28 & - & - & - & - & - & - \\
 & UM.IMAGE.ViT.BERT.BERT & 0.46 & 0.51 & 0.10 & 0.45 & 0.44 & 0.38 & \ul{0.911} \\
 & UM.IMAGE.BEiT.BERT.BERT & 0.40 & 0.50 & 0.11 & 0.44 & 0.44 & 0.38 & 0.909 \\ \midrule
\multirow{13}{*}{MM} & MM.ViT.BERT.BERT & 0.45 & 0.51 & 0.11 & 0.46 & 0.44 & 0.38 & \ul{0.911} \\
 & MM.BEiT.BERT.BERT & 0.44 & 0.48 & 0.09 & 0.45 & 0.45 & 0.39 & 0.910 \\
 & MM-CoT (w/o OCR) & 0.59 & 0.58 & 0.13 & 0.53 & 0.50 & 0.47 & 0.891 \\
 & MM-CoT & 0.67 & \ul{0.59} & 0.12 & \ul{0.54} & \ul{0.51} & \textbf{0.49} & 0.894 \\
 & ViLT & 0.43 & - & - & - & - & - & - \\
 & $\bullet$MM-CoT (w/ Lecture) & \ul{0.69} & \ul{0.59} & 0.13 & \ul{0.54} & \ul{0.51} & \textbf{0.49} & 0.895 \\
 & miniGPT4 (ZS) & 0.32 & 0.09 & 0.00 & 0.14 & 0.21 & 0.23 & 0.753 \\  
 & miniGPT4 (FT) & 0.28 & 0.12 & 0.00 & 0.16 & 0.23 & 0.26 & 0.771 \\
 & LLaVA (ZS) & - &	0.05 &	0.00 &	0.09 &	0.17 &	0.18 &	0.837\\
 & MM-CoT (\texttt{QCML$\rightarrow$A}, w/ LLaVA rationales) & 0.66 & 0.59 & 0.12 & \ul{0.54} & \ul{0.51} & \textbf{0.49} & 0.896 \\\cmidrule(l){2-9}
 & \model\ (w Entity-Specific Rationale) & \textbf{0.87} & 0.58 & 0.17 & 0.53 & \textbf{0.56} & \ul{0.48} & 0.932 \\
 & $\bigstar$\model\ (w Generic Rationale)  & \textbf{0.87} & \textbf{0.63} & \textbf{0.19} & \textbf{0.55} & \textbf{0.56} & 0.46 & \textbf{0.934} \\ \midrule
 \multicolumn{2}{c}{$\Delta_{\bigstar\mbox{--}\bullet}$(\%)} & \textcolor{blue}{18$\uparrow$} &	\textcolor{blue}{4$\uparrow$} &	\textcolor{blue}{4$\uparrow$} &	\textcolor{blue}{1$\uparrow$} &	\textcolor{blue}{5$\uparrow$} &	\textcolor{red}{1$\downarrow$} &	\textcolor{blue}{2$\uparrow$}\\\bottomrule  
\end{tabular}%
}
\caption{Benchmarking results for MemeVQA, comparing the proposed approach vs unimodal and multimodal baselines. Table Footnotes: \textbf{highest}, \ul{second-highest}, $\bullet$: MM-CoT (w Lecture) -- Best Baseline, and $\bigstar$: \model\ (proposed approach). \model\ variants -- \textit{(a). w Entity-Specific:} Utilizes rationale conditioned upon the answer predicted by the first module; and \textit{(b). w Generic:} Utilizes generic rationale.}
\label{tab:primaryobservations}
\end{table*}

\section{Experiments}
In our study, \model\ is rigorously tested against various models, with results averaged over five runs. The \task\ task involves two components: answer prediction and explanation generation, each evaluated using different metrics. Answer prediction is measured for accuracy due to entity imbalance and open-ended nature in the \exhvv\ dataset. Explanation quality is assessed against \exhvv\ ground truth using metrics like BLEU-1, BLEU-4, ROUGE-L, METEOR, CHRF, and BERTScore. Baseline comparisons span uni-modal (text, image) and multi-modal settings. Additionally, \model's robustness is evaluated through diverse question types, confounding-based tests, and multimodal and error analyses.
\begin{figure*}[th!]
    \centering
    \includegraphics[width=\textwidth]{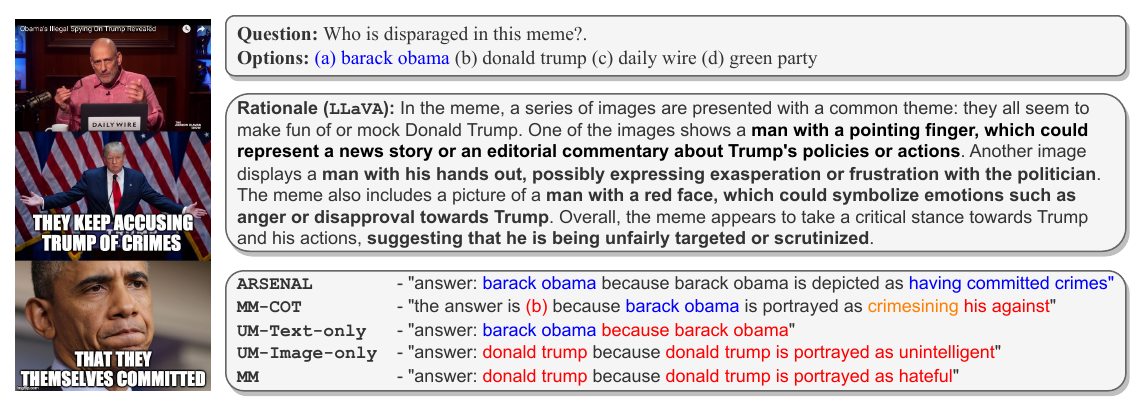}
    \caption{Comparison of \model's output for a sample meme, with four baselines. The LLaVA-based rationale depicted is used for generating the explanation by \model. The font color scheme is as follows: \textcolor{blue}{correct}, \textcolor{red}{incorrect}, and \textcolor{orange}{partially-correct}.}
    \label{fig:observation}
\end{figure*}

\section{Benchmarking \task}
\label{sec:benchmarking}
The T5-based text-only model performs well in answer prediction with an accuracy of $0.53$, outperforming image and multimodal models. However, its explanations are incomplete, repetitive, and lack coherence, resulting in low ROUGE-L ($0.44$), CHRF ($0.35$), and METEOR ($0.41$) scores, second only to the LLM-based miniGPT model.


The ViT model, a strong unimodal image baseline, has low answer prediction accuracy and fluent yet repetitive explanations like the T5 baseline, quantified by low ROUGE-L ($0.45$), METEOR ($0.44$) and CHRF ($0.38$) scores. Both ViT and BEiT unimodal baselines perform poorly, with BEiT scoring $0.40$ accuracy. Multimodal baselines (ViT+BERT, BEiT+BERT) yield answer prediction accuracy ($0.45$ and $0.44$, respectively) similar to that of ViT but slightly outperform the ViT-based unimodal model in terms of the generated explanation qualitatively. This underscores ViT's robustness over BEiT for unimodal and multimodal settings and, consequently, for the Vicuna-based miniGPT4 and LLaVA-based \model.


In the closed-form Visual Question Answering domain, we benchmark against models like the multimodal ViLT, which achieves a fine-tuned accuracy of $0.43$. LLM-based models like miniGPT4 and GPT3.5 show \textit{low} answer prediction accuracy in both zero-shot ($0.32$) and visual description-based fine-tuning ($0.28$) for the former, and $0.28$ for GPT3.5. These models' explanations lack specificity, as indicated by miniGPT4's BLEU-1 score of $0.12$ and ROUGE-L score of $0.16$ post-fine-tuning. Despite their detailed and reasonable reasoning, they fall short in standard evaluations due to their excessive length and nonspecific content. However, BERTScore values of $0.771$ for miniGPT4 (FT) and $0.837$ for LLaVA-based models suggest a reasonable coherence with the memes in question.  


Our primary comparison is to models leveraging the MM-CoT model in various prompt and input settings. The utility of the OCR text is proven by the $8\%$ drop in accuracy observed on eliminating the OCR text from the input. The addition of \textit{generic lectures} (\texttt{L}) also positively impacts the model's performance, with a $2\%$ increment in answer prediction accuracy. Introducing a contextual rationale using zero-shot inferencing using an LLM such as LLaVA presents qualitative improvements in the explanation generation quality of the MM-CoT model.
\begin{table}[t!]
\centering
\resizebox{0.6\columnwidth}{!}{%
\begin{tabular}{@{}ccc@{}}
\toprule
\textbf{Measures} & \textbf{Average} & \textbf{Std. Dev} \\ \midrule
Accuracy          & 0.87             & 0                 \\
BLEU-1            & 0.54             & 0.13              \\
BLEU-4            & 0.15             & 0.06              \\
ROUGE-L           & 0.50             & 0.08              \\
METEOR            & 0.54             & 0.03              \\
CHRF              & 36.63            & 20.29             \\
BERTScore         & 0.92             & 0.02              \\ \bottomrule
\end{tabular}%
}
\caption{Averages and Std. Dev. of \model's performance measured across primary evaluation metrics, over five independent runs.}
\label{tab:reproduce}
\end{table}

It is also worth noting, that the MM-COT model underperforms in understanding memes compared to the new model, which excels in accuracy and explanation due to its \textit{Rationale Generation Module}, offering a deeper contextual grasp of meme content. Table \ref{tab:reproduce} presents the average scores of \model\ across the seven primary metrics explored, over five independent runs.

\paragraph{\textit{Discussion:}}
Our analysis of 60 random test samples compared \model\ with other methods in terms of answer quality, explanation coherence, and modality-specific nuances. \model\, particularly through the LLaVA approach, excels in reasoning and explaining by effectively integrating details from various meme modalities, as shown in Figs. \ref{fig:observation} and \ref{fig:24}. In contrast, the MM-CoT model struggles with syntactic and grammatical correctness (c.f. Figs~\ref{fig:observation}, \ref{fig:3}, and \ref{fig:4}). A T5-based text-only model often produces incoherent and incomplete outputs (c.f. Fig. \ref{fig:3}). The UM.IMG.ViT.BERT.BERT model faces challenges in contextualization and alignment, with explanations that are semantically related yet irrelevant. Image-only approaches and multimodal baselines show a lexical bias, and the MM.ViT.BERT.BERT multimodal setup, despite striving for fluency, fails in complex reasoning, leading to generic explanations (refer to Figs~\ref{fig:observation} and \ref{fig:9}).\footnote{For more details, see App.~\ref{app:multimodal}.} The performance difference might not be as evident from a 2\% quantitative increment observed for a metric like a BERTScore, relative to the 18\% enhancement for answer prediction accuracy but is distinctly visible for the demonstrative example depicted in Fig. \ref{fig:observation}, and Appendix \ref{app:examples}.
\begin{table}[t!]
\centering
\resizebox{\columnwidth}{!}{%
\begin{tabular}{@{}lcccc@{}}
\toprule
\multicolumn{1}{c}{\textbf{Experiment}} & $\mathbf{Q_{div}}$ & \textbf{Yes/No} & \textbf{None (All)} & \textbf{None (Train)} \\ \midrule
UM.TXT.T5 & 0.351 & 0.805 & 0.461 & 0.457 \\
UM.ViT.BERT.BERT & 0.273 & 0.373 & 0.328 & 0.253 \\
MM.ViT.BERT.BERT & 0.341 & 0.295 & 0.474 & 0.438 \\
\model & 0.818 & 0.769 & 0.692 & 0.721\\ \bottomrule
\end{tabular}
}
\caption{Robustness Analysis: (a) Question Diversification ($\mathbf{Q_{div}}$); (b) Confounder Setting (three scenarios).}
\label{tab:confound}
\end{table}


\section{Robustness Analysis}
\label{sec:rob}
A key factor that is expected to characterize the efficacy of a model for a task like \task, is it's robustness to variations within the question/answer formulation. This is also critical due to the resultant variability within the LLM's generated responses \cite{salinas2024butterfly}. To this end, we examine \model's performance in comparison to other contemporary baselines, by (a) Question Diversification, and (b) Confounding Analysis.
\paragraph{Question Diversification:}
In our analysis, we evaluate the performance of \model\ and current baselines using more naturally framed questions than those in \dataset. We achieve question diversity by employing the \texttt{Llama-2-7b-chat} model to generate five unique variations of each original question. Each question is then randomly replaced with one of these generated alternatives, ensuring a wide range of questioning styles.\footnote{Refer App. \ref{app:sec:questdiv} for more details on \textit{Question Diversification}.} 

As an indicator of the robustness of \model\ to diversity in questions, when trained and tested on the new diverse questions, we obtained an answer prediction accuracy of 0.82 (c.f. Table \ref{tab:confound}). This is a marginal decline from its performance of 0.87 on the original setting, having a structured question set. In comparison, the UM.TEXT.T5 baseline descends from an accuracy of 0.53 to 0.35, UM.ViT.BERT.BERT from 0.46 to 0.27 and MM.ViT.BERT.BERT from 0.45 to 0.34. These results are a clear indication that \model\ is able to accommodate significant variations and diversity in the question framing setup while other models are not as robust to these changes.

\paragraph{Confounding Analysis:}
Our study evaluates the robustness of \model\ against contemporary baselines through three confounding settings, crafted to challenge the model with scenarios differing from typical tasks. These settings involve alterations in questions and options. We compare \model\ with three contemporary baseline models: UM.TEXT.T5, UM.IMAGE.ViT.BERT.BERT, and MM.ViT.BERT.BERT. Analyzing \model\ across the \textit{following} settings and against these baselines is crucial for understanding its real-world applicability and performance. For detailed information on these confounding tasks, see App. \ref{app:confounder}.

\paragraph{\textit{Confounder A -- Yes/No Confounding:}}
Transforming dataset to binary `yes or no' questions ($50\%$ chance), reshaping `yes' as ``Is [answer] [rephrased question]?'' and altering `no' by modifying role labels.

\paragraph{\textit{Confounder B -- None Sampling Across All Sets:}}
Replacing $20\%$ of answers with `None' by swapping role labels, maintaining consistency; $M_{new} = \{M, None\}$ across sets.

\paragraph{\textit{Confounder C -- None Sampling Across Train Only:}}
Introducing $20\%$ random `None' answers in training; model adapts to `None' while testing remains unchanged; $M_{new} = \{M, None\}$ across sets.


The 'yes or no' confounding setting in our study allows for assessing the model's reasoning robustness. Models depending on statistical probabilities fail here, as answers can be paired with either correct or incorrect role labels regardless of their dataset frequency. \model\ and UM.TEXT.T5 demonstrate strong reasoning skills, with scores of $0.77$ and $0.80$ respectively, indicating they rely on reasoning over statistics. In contrast, the UM.IMAGE.ViT.BERT.BERT-based model and MM.ViT.BERT.BERT-based model score poorly at $0.37$ and $0.29$, highlighting their reliance on statistical likelihoods of answers based on dataset frequency.

We also evaluated the robustness and generalizability of \model, using two settings involving ``None'' answers. Notably, only \model\ delivers good performance ($0.69$ accuracy) in the more challenging original testing set compared to the revised set, showcasing its better generalizability despite being trained on valid ``None'' answers data.

\section{Conclusion and Future Work}

This study introduced \task, a task that involves multimodal question answering for image-text memes, delving into their intricate visual and linguistic layers. Utilizing recently open-sourced LLMs, especially their multimodal adaptations, we tackled the challenge of complex, non-trivial multimodal content. Through a new dataset, \dataset, we assessed systems' reasoning in assigning semantic roles to meme entities via \textit{question-answering} and \textit{contextualization} based objectives. Our experiments showcased the efficacy of the proposed two-stage training framework, \model, while leveraging  existing language models and multimodal LLMs, to outperform the state-of-the-art by a remarkable $18\%$ accuracy gain. This study reveals the potency and limitations of multimodal LLMs, enabling the scope for sophisticated setups embracing diverse questions, domains, and emotional nuances conveyed through memes. Ultimately, our findings steers future exploration and the development of comprehensive systems dedicated to deciphering memetic phenomena.

Our future aim is to create sophisticated, multi-perspective sets for \task, moving beyond standard QnA towards an optimal multimodal solution.


\section*{Acknowledgements}
The work was supported by Wipro research grant.

\section*{Limitations}

This section highlights \model's limitations, including semantically inconsistent rationales, factual errors, and multimodal bias, inherent to LLaVA's generation capacity. For some cases, LLaVA's rationales seem to be mining the inductive biases due to the co-occurrences of disparate keywords while being influenced by LLM's pre-training corpus and web data, exhibited mostly for \textit{missing-modality} and high \textit{inter-modal incongruity}. An example for the latter shown in Fig. \ref{fig:error1} (c.f. Appendix \ref{app:error}) illustrates how biased inference by LLaVA dilutes \model's output due to inaccurate contextualization, whereas MM-CoT deduces the answer accurately, possibly due to standardized definitions being used instead of LLM-based rationales.


\section*{Ethics and Broader Impact}
\label{sec:ethics}

\paragraph{Reproducibility.}  We present detailed hyper-parameter configurations in Appendix~\ref{app:sec:hyp} and Table~\ref{tab:hyper}. 

\paragraph{User Privacy.} 
The information depicted/used does not include any personal information. 


\paragraph{Biases.}
Any biases found in the source dataset \exhvv\ are attributed to the original authors \cite{Sharma_Agarwal_Suresh_Nakov_Akhtar_Chakraborty_2023}, while the ones in the newly constructed dataset is unintentional, and we do not intend to cause harm to any group or individual. 


\paragraph{Misuse Potential.}
The ability to identify implied references in a meme could enable wrongdoers to subtly express harmful sentiments towards a social group. By doing this, they aim to deceive regulatory moderators, possibly using a system similar to the one described in this study. As a result, these cleverly crafted memes, designed to carry harmful references, might escape detection, thereby obstructing the moderation process. To counteract this, it is advised to incorporate human moderation and expert oversight in such applications.

\paragraph{Intended Use.}
We make use of the existing dataset in our work in line with the intended usage prescribed by its creators and solely for research purposes. This applies in its entirety to its further usage as well. We do not claim any rights to the dataset used or any part thereof. We believe that it represents a useful resource when used appropriately.

\paragraph{Environmental Impact.}

Finally, large-scale models require a lot of computations, which contribute to global warming \cite{strubell2019energy}. However, in our case, we do not train such models from scratch; instead, we fine-tune them on a relatively small dataset.

\bibliography{ARRDec23/CommonFiles/custom}

\clearpage
\appendix

\section{Hyper-parameter and Implementation}
\label{app:sec:hyp}
We train all the models using PyTorch on an actively dedicated NVIDIA Tesla V100 GPU, with 32 GB dedicated memory, CUDA-12.2 and cuDNN-7.6.5 installed. For all the models with the exclusion of LLaVA and MiniGPT4, we import all the pre-trained weights from the \texttt{huggingface}\footnote{\url{https://huggingface.co/models}} API. Additionally, we used a series of architectural additions and delta weights to obtain \texttt{LLaVA-7B-v0}\footnote{\url{https://github.com/haotian-liu/LLaVA}} from the base \texttt{LLaMA-7B} model available under an academic license from Meta. We randomly initialize the remaining weights.

\begin{table}[t!]
    \centering
    \resizebox{\columnwidth}{!}{%
    \begin{tabular}{ccccc}
    \toprule
    \textbf{Modality} & \textbf{Model} & \textbf{LR} & \textbf{BS} & \textbf{\# Params (M)} \\
    \midrule
    \multicolumn{1}{c}{\multirow{3}{*}{UM}} & TEXT T5 & 1.00E-4 & 4 & 222.9 \\
    & IMG ViT-BERT & 5.00E-5 & 4 & 333.7 \\
    & IMG BEiT-BERT & 5.00E-5 & 4 & 333.0 \\
    \midrule
     \multicolumn{1}{c}{\multirow{8}{*}{MM}} & ViT-BERT & 5.00E-5 & 4 & 333.7 \\
    & BEiT-BERT & 5.00E-5 & 4 & 333.0 \\   
    & ViLT & 5.00E-5 & 8 & 113.4 \\
    & MM-CoT (\texttt{allenai-t5-base}) & 5.00E-5 & 4 & 226.6\\
    & MM-CoT (\texttt{t5-large}) & 5.00E-5 & 4 & 744.2 \\
    & MM-CoT (\texttt{allenai-t5-large}) & 5.00E-5 & 4 & 744.2 \\
    & \model\ - Answer Prediction & 5.00E-5 & 4 & 744.2\\
    & \model\ - Explanation Generation & 1.00E-4 & 4 & 737.6\\
    \bottomrule   
    
    \end{tabular}
    }
    \caption{Hyper-parameters}
    \label{tab:hyper}
\end{table}
Most of our models are implemented using the Adam optimiser \cite{kingma2014adam} with a learning rates as specified in Table \ref{tab:hyper}, a weight decay of $1e^{-5}$. We use a Cross-Entropy (CE) and a language modeling loss (LML) as per the applicability. We conducted a thorough empirical analysis before freezing the optimal set of hyperparameters for the current task for all the models examined. We also early stop to preserve our best state convergence for each experiment. Further details of hyperparameters employed can be referred to from Table~\ref{tab:hyper}. On average, it took approx. 2:30 hours to train a typical multimodal neural model on a dedicated GPU system. 

For \model,\ we use a learning rate of $1e^{-4}$, with \texttt{eps}=($1e^{-30}$, $1e^{-3}$), \texttt{clip\_threshold}=1.0, \texttt{decay\_rate}=-0.8, \texttt{weight\_decay}=0.0. Moreover, we set the \texttt{max\_source\_length} = 512 and \texttt{max\_target\_length} = 256 in first-step, QCM-LE task of MM-CoT, \texttt{max\_target\_length} = 16 in the second-step QCMG-A task of MM-CoT, and \texttt{max\_target\_length} = 32 in the explanation generation module using T5-Large.

\begin{figure}[t!]
    \centering
    \includegraphics[width=\columnwidth]{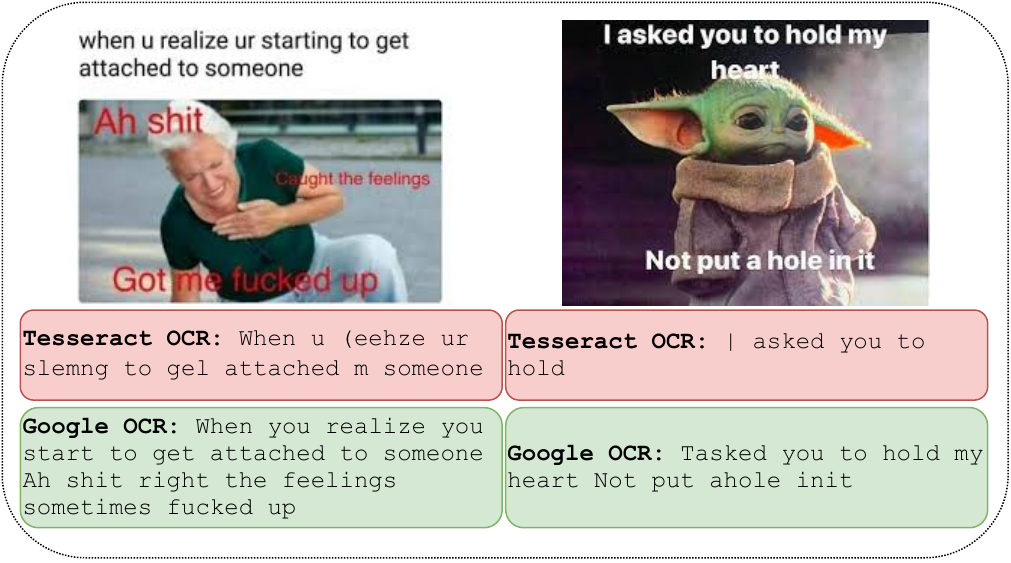}
    \caption{Comparison b/w the quality of the OCR-extracted text via (a) Tesseract OCR, and (b) Google OCR.}
    \label{fig:ocr_comp}
\end{figure}

\paragraph{A note on \textit{fine-tuning}:} Fig. \ref{fig:enter-label} illustrates that the \textit{Answer Prediction Modules} and \textit{Explanation Generation Module} are fine-tuned (please refer to the symbol legend at the top-right corner of Fig. \ref{fig:enter-label}) components within the proposed framework. The \textit{Answer Prediction Module} implementation follows guidelines from the multimodal question-answering with CoT study \cite{zhang2023multimodal}. The LLaVa model generates a rationale that is utilized as a proxy for the intermediate rationale for MM-CoT. Furthermore, the \textit{Explanation Generation Module} is fine-tuned using the T5-Large model, employing a training approach akin to that used for the UM\_T5 model, also detailed in the implementation of the source code.

\section{Text Extraction via OCR}
\label{app:ocr}
The OCR data is part of the original ExHVV dataset, as released with the original work \cite{Sharma_Agarwal_Suresh_Nakov_Akhtar_Chakraborty_2023}, which was extracted using Google GCV OCR\footnote{\href{https://cloud.google.com/vision/docs/ocr}{Google Cloud Vision OCR API}} (GOCR) as primary inputs. The OCR for each meme is available as part of ExHVV, and we have not made any modifications to this data field. Text retrieval through \textit{optical character recognition} (OCR) is crucial for extracting text from memes. The efficacy of the OCR method impacts the system's overall performance. Towards examining the quality of the GOCR approach used originally, we examine the text retrieval capabilities of \textit{two} widely-used OCR-based APIs: Google Tesseract API\footnote{\href{https://pypi.org/project/pytesseract/}{Google's Tesseract-OCR API}} (TOCR) and Google GCV API (GOCR), for this purpose.

Our qualitative assessment of 30 varied memes reveals occasional errors in TOCR and fewer in GOCR. TOCR errors are frequent in challenging scenarios, such as overlapping text and images, low-quality graphics, or small text. In contrast, GOCR often outperforms TOCR, even in simpler situations. Figure \ref{fig:ocr_comp} illustrates the disparity in text extraction accuracy between TOCR and GOCR. The first example in Fig. \ref{fig:ocr_comp} (\textit{left}) shows a combination of straightforward and complex elements like clear black text on a white background and intricate visual-text overlaps, where TOCR fails but GOCR succeeds. Conversely, the second example in Fig. \ref{fig:ocr_comp} (\textit{right}), a simpler meme, presents more difficulties for TOCR, while GOCR maintains clarity.

\section{Motivation for \model's Design} 
Within the context of reasoning-based question-answering setup for memes, relevant solutions are scarce within the realm of neural frameworks and multimodal LLMs as we transition between them. For existing multimodal-LLM-based systems, eliciting relevant answers and generating concise explanations for memes is challenging. Strategies would typically solicit systematic instruction-tuning for fine-grained meme-related use cases instead of typical vision+language tasks \cite{liu2023visual,zhu2023minigpt4,zhao2023chatbridge}, which itself has been an active research domain. An alternate solution would be to improve existing multimodal-neural frameworks \cite{zhang2023multimodal} that perform reasoning-based question answering, albeit with constrained reasoning capacity and generative coherence for memes. In this work, we primarily focus on the latter while assessing the potential and limitations of other contemporary solutions.

The capability of the proposed approach (ARSENAL) towards addressing the nuanced complexity posed by memetic content stems from the choice of leveraging detailed rationales generated via multimodal LLM’s, while adapting conventional approaches involving chain-of-thought reasoning which we found in our study are more suited for more accurate answer prediction and focused explanation generation.

\section{More on Prompting Configuration Analysis}
\label{app:prompt}

Using the \texttt{allenai/unifiedqa-t5-base}-based MM-CoT setup, we first evaluate the optimal ordering of lecture (\texttt{L}), explanation (\texttt{E}), and answer (\texttt{A}) components for \task. Comparing \texttt{LEA} and \texttt{ALE} configurations, we find a significant $22\%$ accuracy difference, emphasizing ordering importance. The two-stage setup generally outperforms one-stage, except for \texttt{QCML$\rightarrow$A} (AT5B), suggesting optimal answer inference with lecture/explanation-based reasoning. The first-stage training with rationale/explanation benefits \texttt{QCM$\rightarrow$LE} and \texttt{QCML$\rightarrow$E} configurations. Among three language models (\texttt{unifiedqa-t5-base/large}, \texttt{t5-large}), the two-stage \texttt{t5-large} achieves the highest accuracy of $0.776$, slightly better than \texttt{unifiedqa-t5-large}.

It is also worth noting that the accuracy of the two-stage framework, with configuration [\texttt{QCM$\rightarrow$L}, \texttt{QCMG$\rightarrow$AE}] using \texttt{allenai/unifiedqa-t5-base} comes out to be $1.5\%$ higher than that from \texttt{t5-large}. This could be attributed to the format-agnostic design of the former, the efficacy of which could be best seen for the challenging \texttt{*$\rightarrow$AE}-based scenarios in the two-stage setup (see Figure~\ref{fig:prompt}). In addition to this, performing inference with \texttt{AE} as outputs mainly yields poor results, as can be observed for the experiments with configurations as \texttt{QCML$\rightarrow$AE} (AT5B), [\texttt{QCM$\rightarrow$L}, \texttt{QCMG$\rightarrow$AE}] (T5L), and [\texttt{QCM$\rightarrow$L}, \texttt{QCMG$\rightarrow$AE}] (AT5B), on average yielding an accuracy of $0.47$. This could be due to the distributional differences between the answer choices and explanations, which the MM-CoT-based setup is unable to adjudicate as part of modeling.

\paragraph{A high-level overview of \textit{prompting scenarios}:}
Our experiments utilize prompting across \textit{three} distinct scenarios and configurations. Sec. \ref{sec:proposed_approach} addresses the prompting setups for the Multimodal CoT model within the answer prediction module. The prompting structure, as explained in the paragraph on \textit{prompting configurations} in Sec \ref{sec:proposed_approach}, follows an \texttt{input$\rightarrow$output} format, with both the input and output comprising combinations of elements denoted by \texttt{QCMLEAG}. Here, \texttt{Q} stands for Question, \texttt{C} for Context, \texttt{M} for multiple options, \texttt{L} for lecture, \texttt{E} for explanation, \texttt{A} for answer, and \texttt{G} for generated intermediate text. In the ARSENAL framework, a two-stage setup is implemented, with prompts formatted as \texttt{QCM$\rightarrow$LE} initially, then \texttt{QCMG$\rightarrow$A}. An illustrative example is provided below: \texttt{QCM} - ``Question: What is slandered in this meme? \textbackslash n Context: \{ocr text\} \textbackslash n Options: (a) antifa (b) democratic party (c) black community (d) conservatives'' \texttt{LE} - ``Solution: \{lecture = generic rationale, R\_generic\} \{explanation\}'' \texttt{QCMG} - ``Question: What is slandered in this meme? \textbackslash n Context: \{ocr text\} \textbackslash n Options: (a) antifa (b) democratic party (c) black community (d) conservatives \textbackslash n \{generated text\}'' \texttt{A} - ``The answer is (a)''. The input for the explanation generation module is detailed in the description leading upto the equation \# \ref{eq:5}, as `Summarize the explanation for \{question\} based on the \{answer\}. Explanation: \{R\_specific or entity-specific rationale\}' Additionally, Sec. \ref{app:sec:questdiv} elaborately discusses the prompt setups used for \textit{question diversification}.

\section{Multimodal Analysis of \model}
\label{app:multimodal}
\begin{figure*}[th!]
    \centering
    \includegraphics[width=\textwidth]{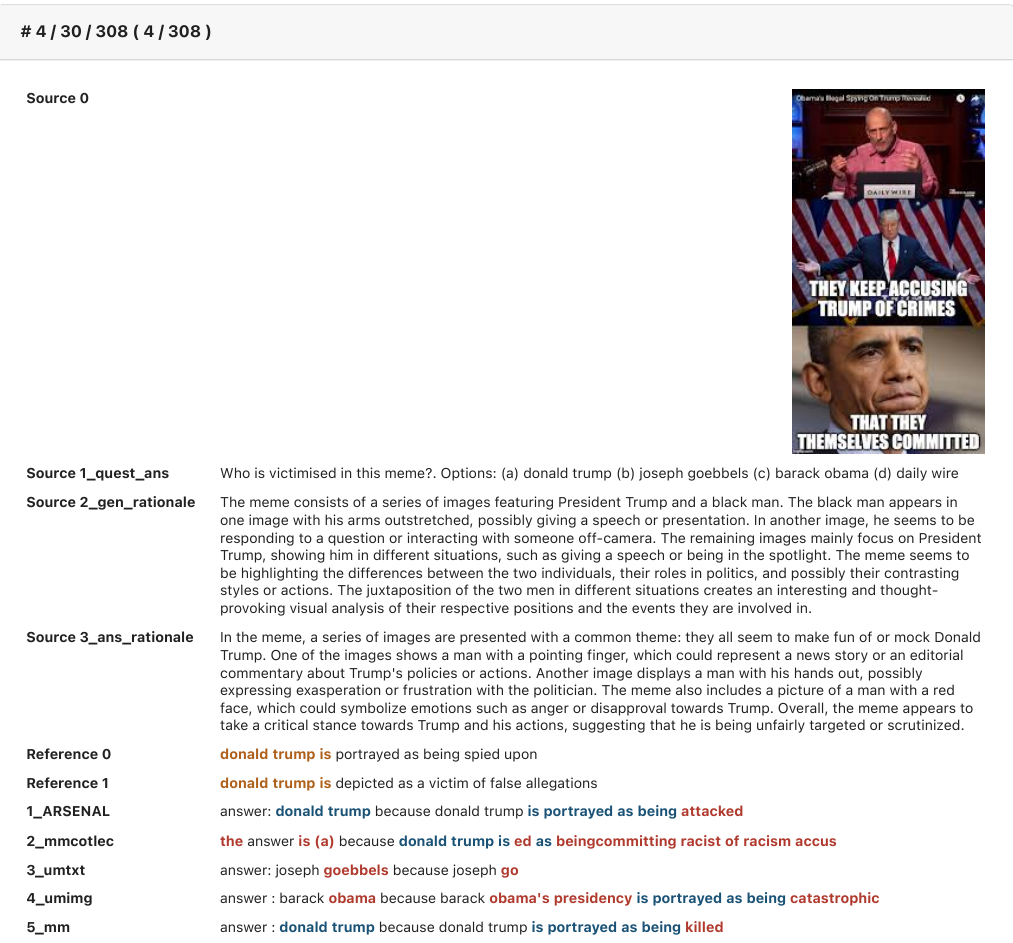}
    \caption{Example 4}
    \label{fig:4}
\end{figure*}

\begin{figure*}[th!]
    \centering
    \includegraphics[width=\textwidth]{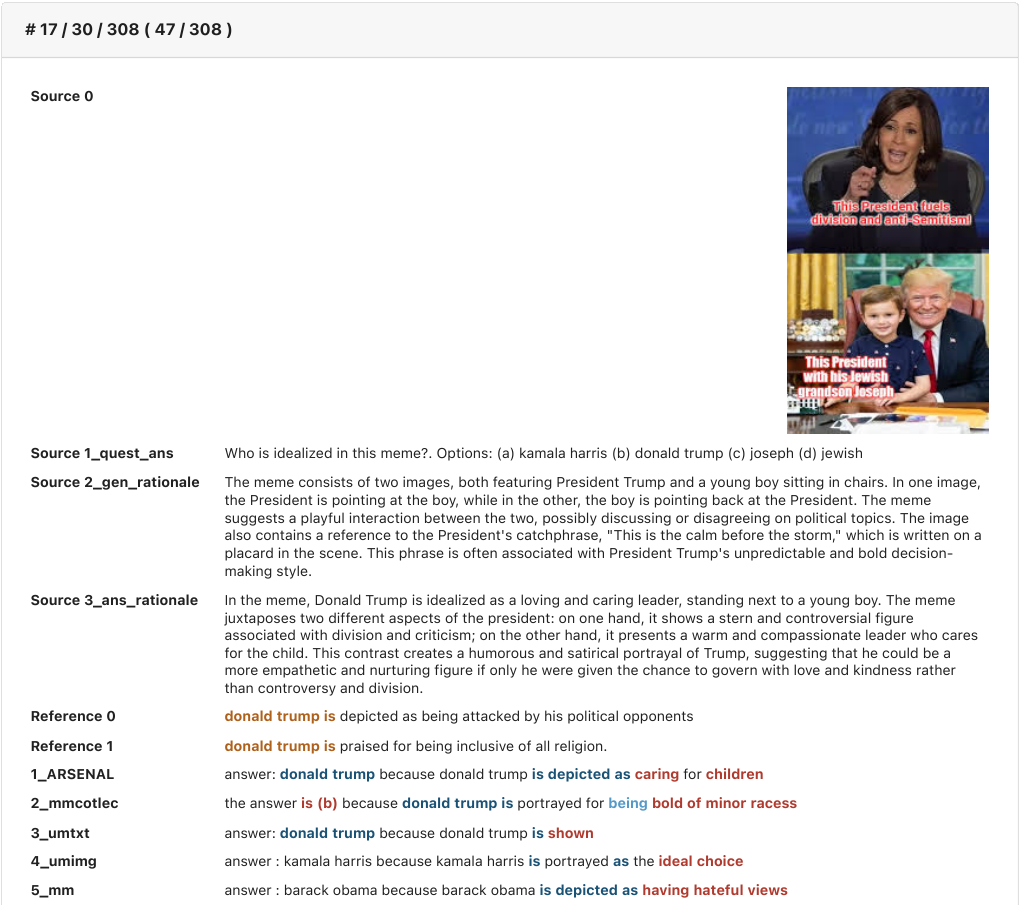}
    \caption{Example 47}
    \label{fig:47}
\end{figure*}

\begin{figure*}[th!]
    \centering
    \includegraphics[width=\textwidth]{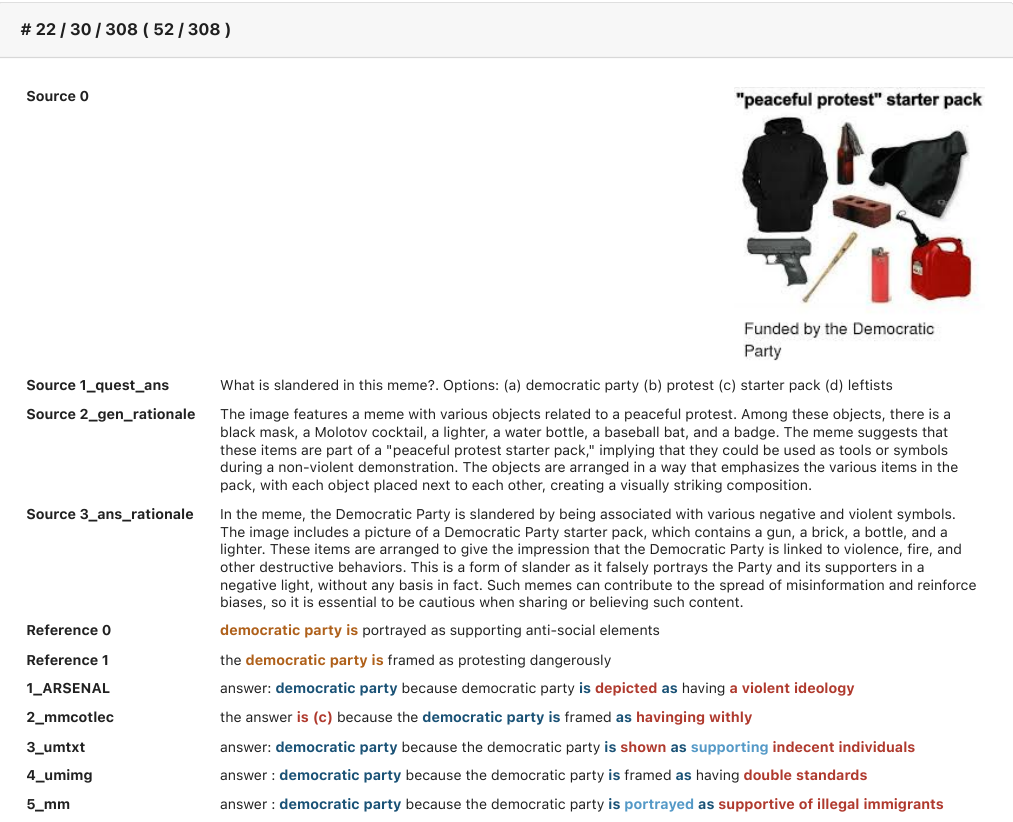}
    \caption{Example 52}
    \label{fig:52}
\end{figure*}

\begin{figure*}
    \centering
    \includegraphics[width=\textwidth]{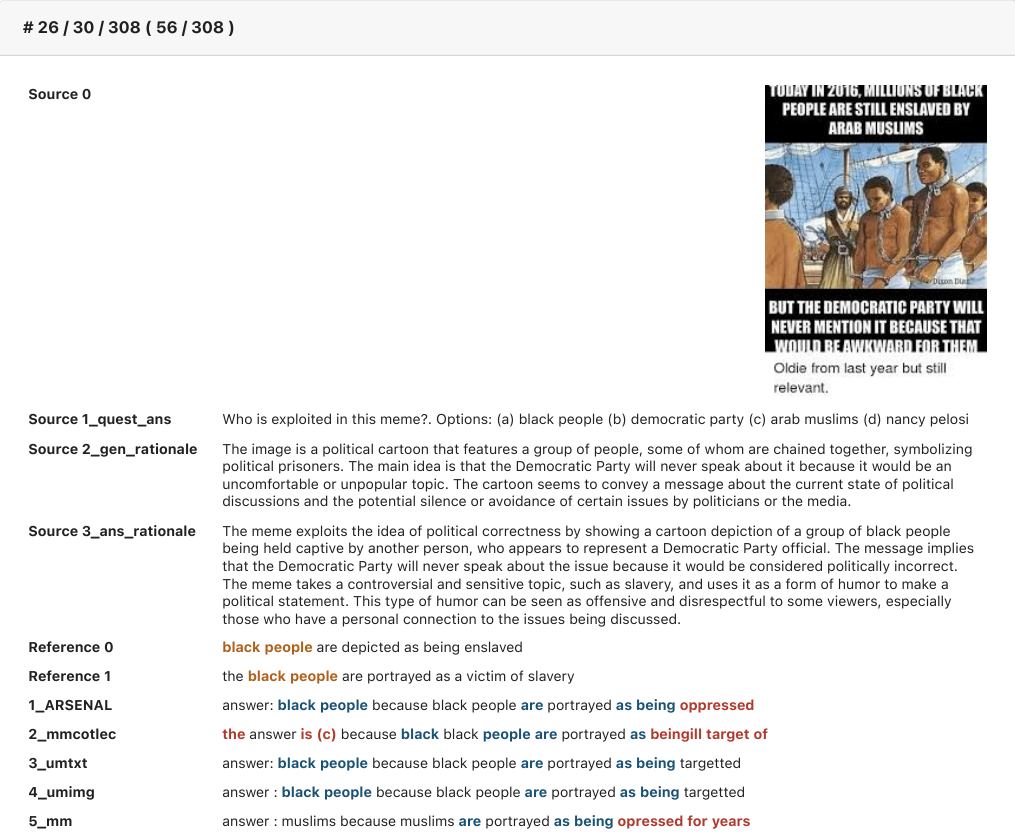}
    \caption{Example 56}
    \label{fig:56}
\end{figure*}

\begin{figure*}
    \centering
    \includegraphics[width=\textwidth]{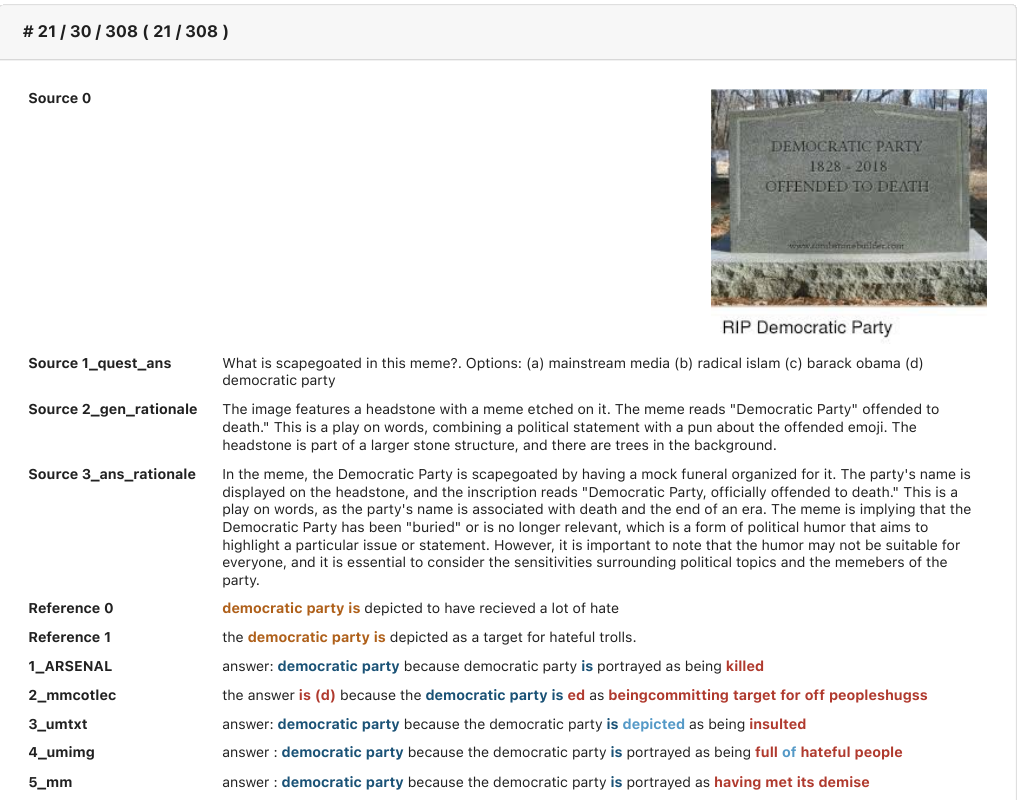}
    \caption{Example 21}
    \label{fig:21}
\end{figure*}

\begin{figure*}
    \centering
    \includegraphics[width=\textwidth]{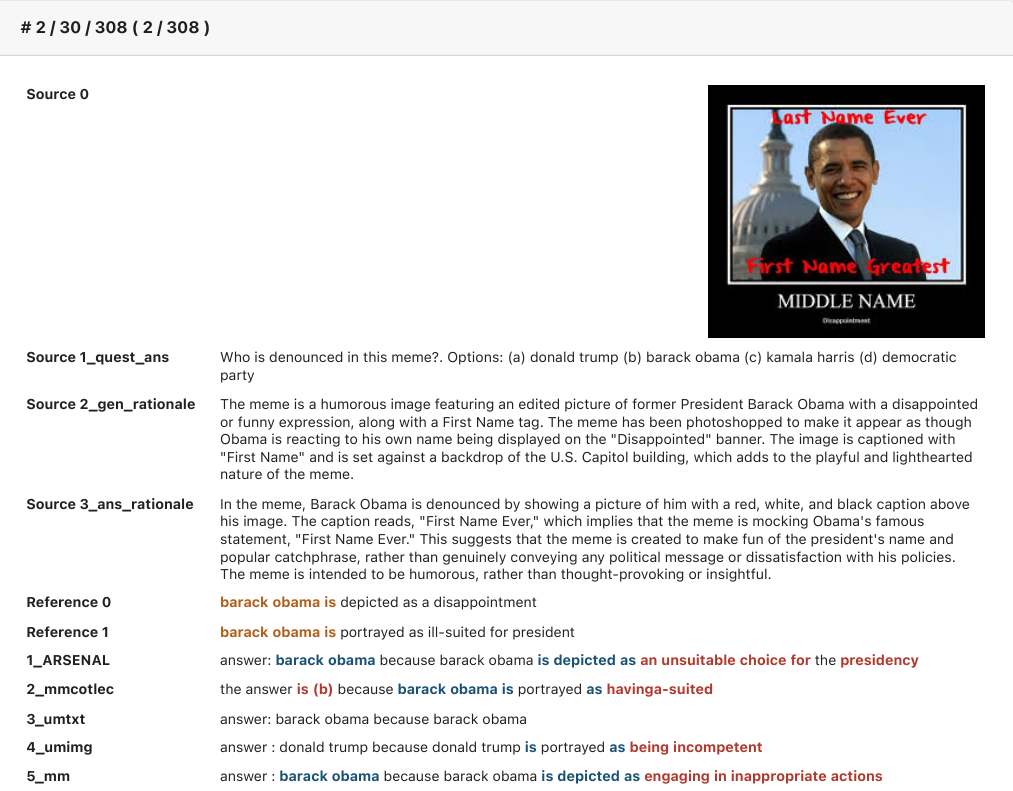}
    \caption{Example 2}
    \label{fig:2}
\end{figure*}

\begin{figure*}
    \centering
    \includegraphics[width=\textwidth]{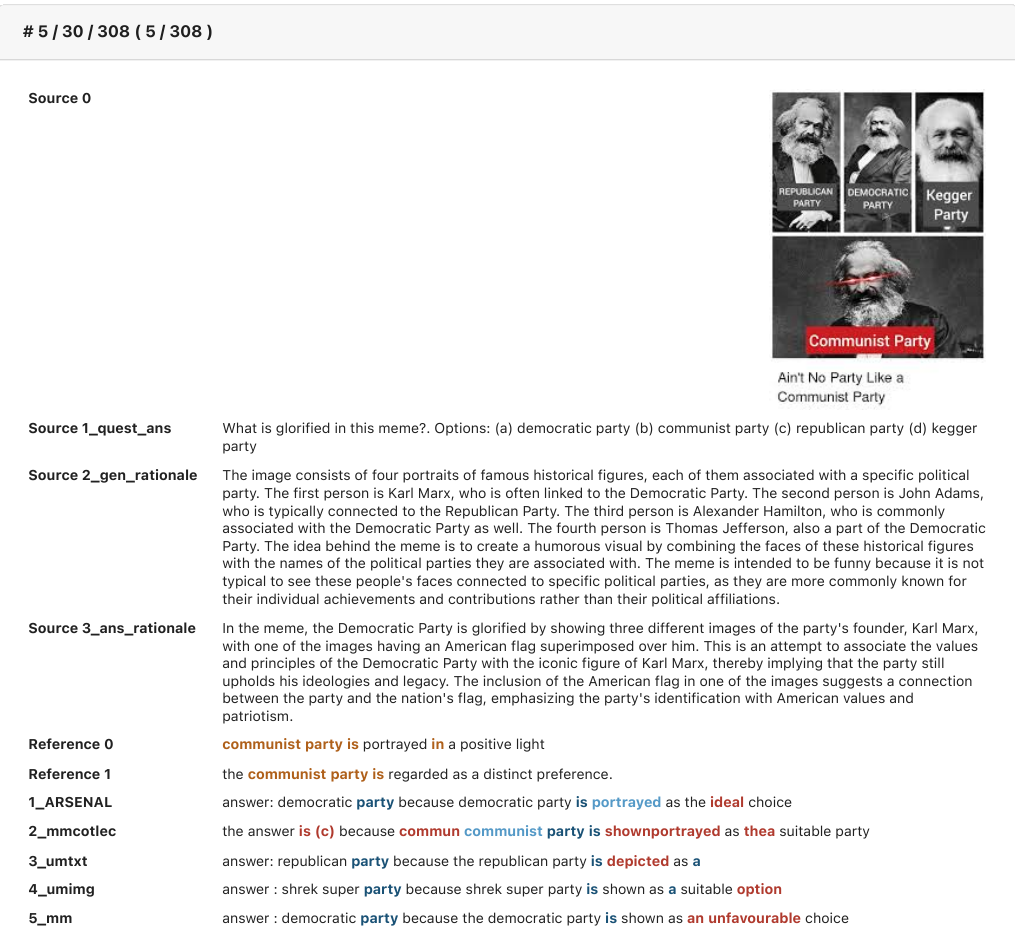}
    \caption{Example 5}
    \label{fig:5}
\end{figure*}

\begin{figure*}[th!]
    \centering
    \includegraphics[width=\textwidth]{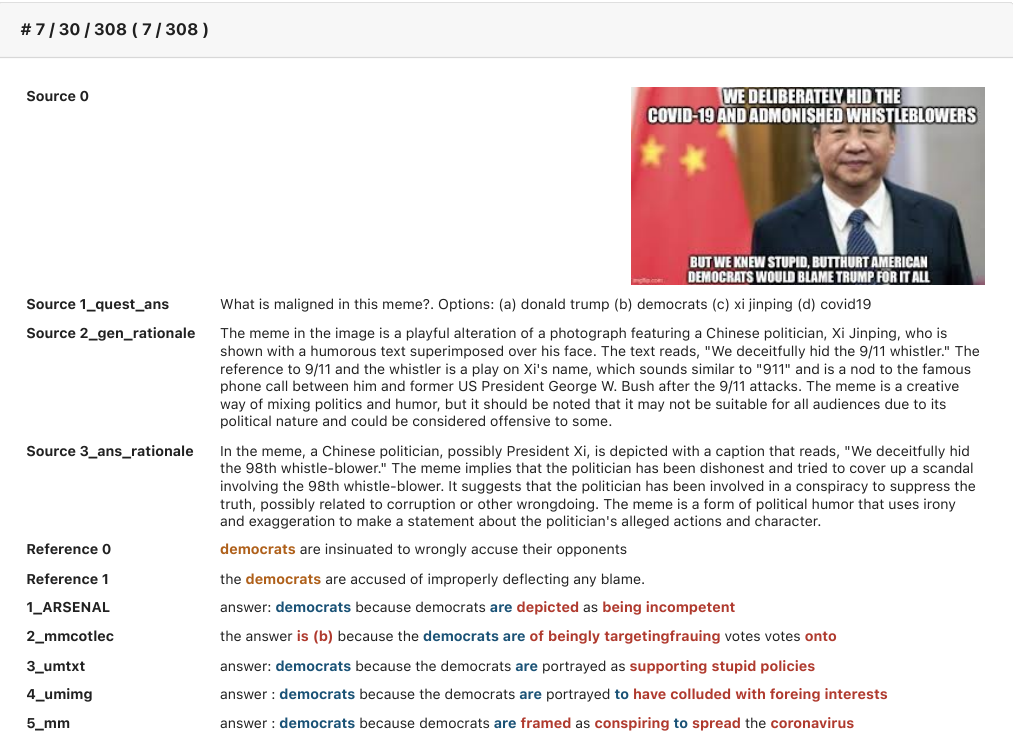}
    \caption{Example 7}
    \label{fig:7}
\end{figure*}

\begin{figure*}
    \centering
    \includegraphics[width=\textwidth]{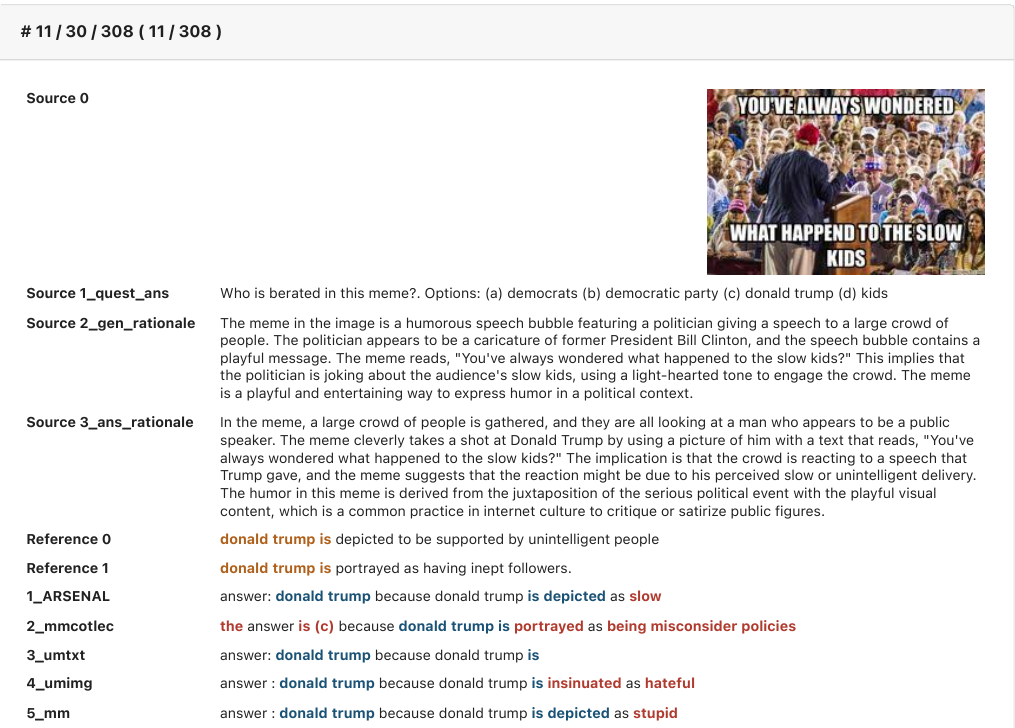}
    \caption{Example 11}
    \label{fig:11}
\end{figure*}

\begin{figure*}
    \centering
    \includegraphics[width=\textwidth]{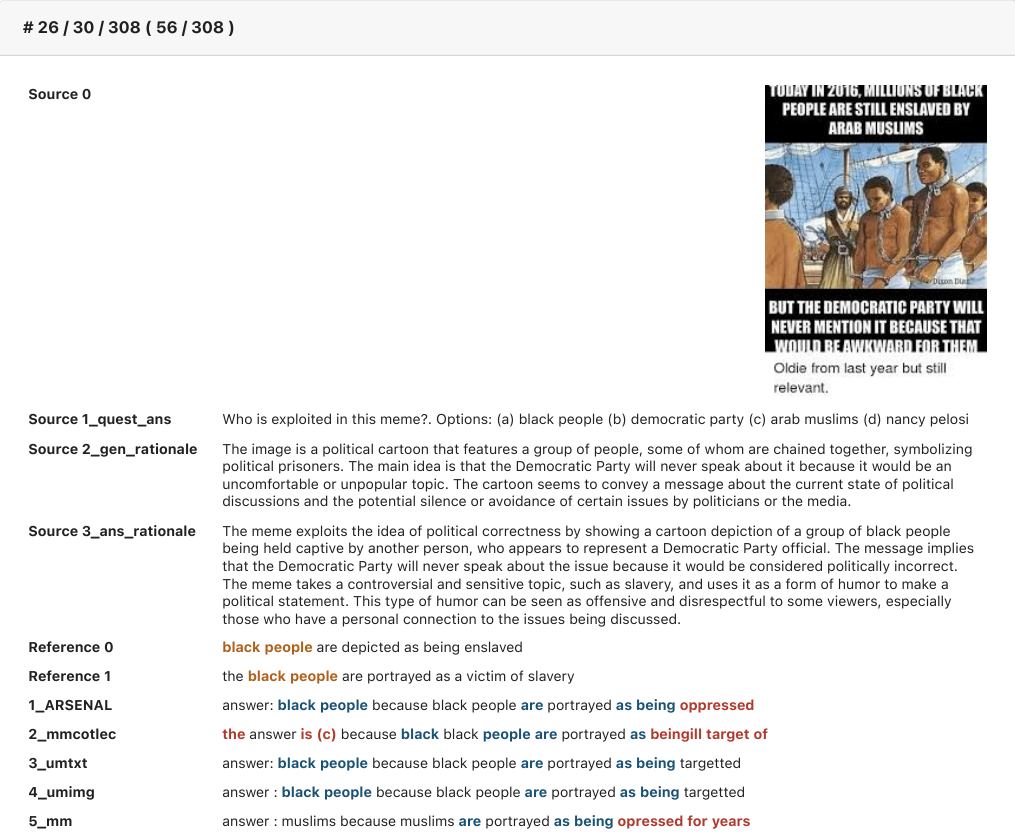}
    \caption{Example 45}
    \label{fig:45}
\end{figure*}

\begin{figure*}
    \centering
    \includegraphics[width=\textwidth]{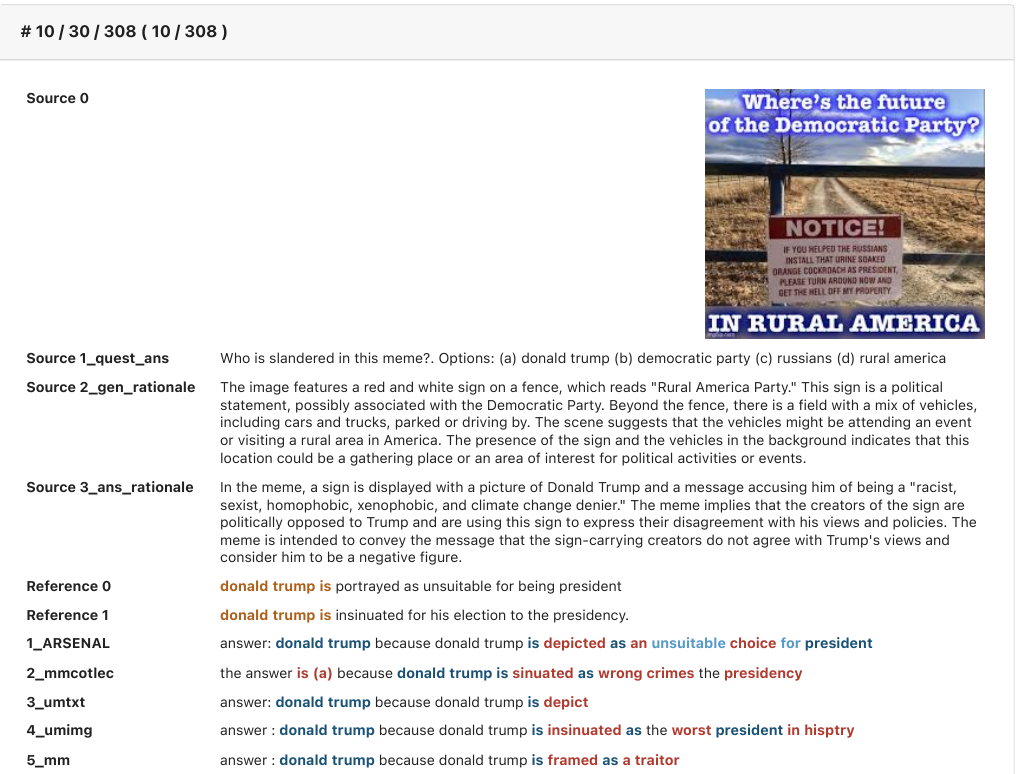}
    \caption{Example 10}
    \label{fig:10}
\end{figure*}

Cross-modal reasoning is a pivotal aspect of LLaVA's capability, particularly evident in situations where textual information falls short. Impressively, LLaVA harnesses its adeptness in detailed visual assessment and intricate reasoning, leading to the generation of semantically accurate rationales, as depicted in Fig. \ref{fig:47}, \ref{fig:52}, and \ref{fig:56}. However, the landscape of cross-modal noise, demonstrated by the example in Fig. \ref{fig:21}, introduces an intriguing challenge. This pertains to cases like \textit{visual exaggeration}, where multimodal models tend to anchor their explanations across multiple modalities without a clear emphasis on a primary one, which could otherwise be self-explanatory. On a related note, the phenomenon of \textit{multimodal hallucinations}, represented by Fig. \ref{fig:2}, \ref{fig:4}, \ref{fig:5}, \ref{fig:7}, \ref{fig:11}, and \ref{fig:45}, brings about an intriguing facet of LLaVA's capabilities. In these instances, the model's explanations may indeed prove accurate, despite the rationales not always aligning with factual accuracy. Such discrepancies might arise due to extrapolated ideas or statements, as well as visual misinterpretation, yet these rationales consistently maintain a high degree of semantic relevance, an observation supported by Fig. \ref{fig:4} and \ref{fig:10}. In light of these intriguing insights, multimodal analysis error analysis emerges as a critical component for understanding LLaVA's performance and refining its cross-modal reasoning and explanation generation abilities.
\begin{figure*}[th!]
    \centering
    \includegraphics[width=\textwidth]{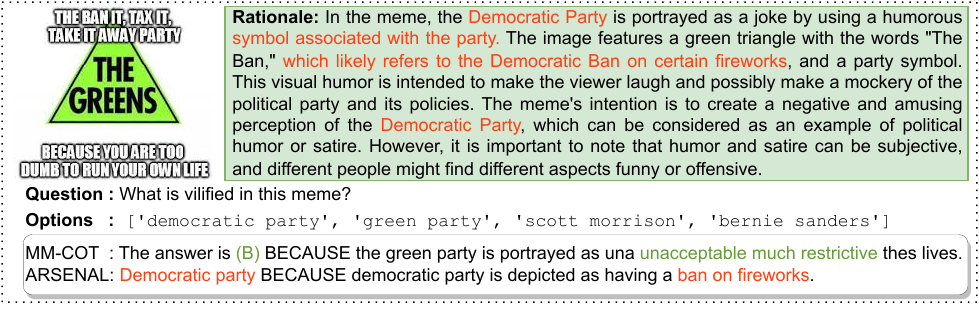}
    \caption{An example of the error-type committed by \model\ (proposed approach) vs. the correct inferencing by the MM-COT based approach.}
    \label{fig:error1}
\end{figure*}

\section{Difference with MM-CoT framework}
The original MM-COT model, while being a strong comparative baseline, lags behind the proposed model, both in terms of answer prediction accuracy and explanation generation quality (18\%-Accuracy and 2\%-BERTScore performance difference w.r.t. ARSENAL), because of its inability to interact and reason well w.r.t. Visual-linguistic semantics of memes. Memes require a deeper understanding of the humour, sarcasm, and hidden meaning of the content, which the MM-COT model is observed to fall short of. The introduction of the Rationale Generation Module is a major contributing factor in the performance of the proposed framework as it provides deeper contextual information about the meme.

\section{Comparison with GPT 3.5 and GPT4} 
\label{comparisonOAI}
As a proxy for comparison with the closed and commercial models like GPT-3.5 and GPT-4, we have provided a comparison with open-source multimodal LLM alternatives in Table 2 in the form of a comparison with LLaVA and miniGPT4 (in zero-shot and fine-tuned settings). The primary reason for this comparison was the accessibility of the technical and background details of these systems in the public domain and to encourage healthy competition within open-source community, especially considering their impressive performance on various multimodal tasks like miniGPT4 exhibiting various emerging capabilities \cite{liu2023llava} and LlaVA achieving SOTA on 11 benchmarks \cite{zhu2023minigpt4}, with rarely any in-depth study w.r.t content like memes, which have very strong visual-linguistic incongruity, in contrast to typically visual-linguistic grounding tasks and datasets.

\begin{table}[t!]
\centering
\resizebox{\columnwidth}{!}{%
\begin{tabular}{@{}lccccc@{}}
\toprule
\multicolumn{1}{c}{\textbf{Approaches}} & \textbf{WER} & \textbf{MEL} & \textbf{WIL} & \textbf{WIP} & \textbf{CER} \\ \midrule
\model & 0.60 & 0.57 & 0.77 & 0.23 & 0.41 \\
MM-CoT (w Lecture) & 0.37 & 0.37 & 0.58 & 0.42 & 0.31 \\
UM.TEXT.T5 & 0.67 & 0.65 & 0.82 & 0.18 & 0.53 \\
UM.IMAGE.BEiT.BERT.BERT & 0.90 & 0.81 & 0.95 & 0.05 & 0.60 \\
MM.ViT.BERT.BERT & 0.89 & 0.81 & 0.95 & 0.05 & 0.60 \\ \bottomrule
\end{tabular}%
}
\caption{Error rate comparison between \model, MM-CoT, unimodal (image and text), and multimodal baselines.}
\label{tab:errorrates}
\end{table}

\section{A note on Ablation Study}

Our ablation analysis begins with a detailed discussion on the investigating \textit{Prompting Configuration} (c.f. Sec. \ref{sec:proposed_approach}, second paragraph, and Fig. \ref{fig:prompt}), and is then reflected as part of \textit{Benchmarking \model} (c.f. Sec. \ref{sec:benchmarking}, and Table \ref{tab:primaryobservations}). The specific experiments reflecting the ablation results are outlined below:

\paragraph{Prompting Configuration (c.f. Fig. \ref{fig:prompt}):} We have explored various permutations of the elements denoted by the acronym \texttt{QCMLEAG} (Question \texttt{Q}, Context \texttt{C}, Multiple Options \texttt{M}, Lecture \texttt{L}, Explanation \texttt{E}, Answer \texttt{A}, and Generated Intermediate Text \texttt{G}). These elements are crucial to the task and solution framework proposed (\model), with the goal of identifying the most effective input-output configurations for the foundational multimodal framework. These experimental explorations were carried out using initial lectures (excluding the more complex LlaVa-based justifications).

\paragraph{Benchmarking MemeMQA (c.f. Table \ref{tab:primaryobservations}):} The experiments labeled under ``Model'' entries such as MM-CoT (without OCR), MM-CoT, MM-CoT (with Lecture), MM-CoT (\texttt{QCML$\rightarrow$A}, with LLaVA rationales), \model\ (with Entity-Specific Rationale), and \model\ (with Generic Rationale), collectively contribute to the ablation analysis for \model. These experiments cover both the basic MM-CoT frameworks and the evolving ARSENAL variants, leading up to the solution ultimately proposed.

\section{Error Analysis}
\label{app:error}
Among various errors in \model's outputs, we found errors due to (a) semantically inconsistent rationales caused by LLaVA, (b) factually incorrect rationales, and (c) multimodal bias. \textit{Semantically inconsistent rationales} are prominent when high inter-modal incongruity occurs. Illustrated in Fig. \ref{fig:error1} (c.f. Appendix \ref{app:error}), a biased inference towards the `democratic party' by LLaVA leads to incorrect predictions in \model. Despite a \textit{green triangle} and the term \textit{party} in the meme, the model lacks cues to understand context. It seems to capture inductive biases from the co-occurrence of `party' and `ban', likely influenced by media coverage and the LLM's training. Whereas, MM-CoT approach accurately predicts the meme's answer and produces somewhat aligned explanations. This is achieved through standardized definitions replacing rationales, aiding the T5 model's inference to connect visual elements and text to the second option.\footnote{For more error-type details, see Appendix. \ref{app:error}.}

\begin{figure*}
    \centering
    \includegraphics[width=\textwidth]{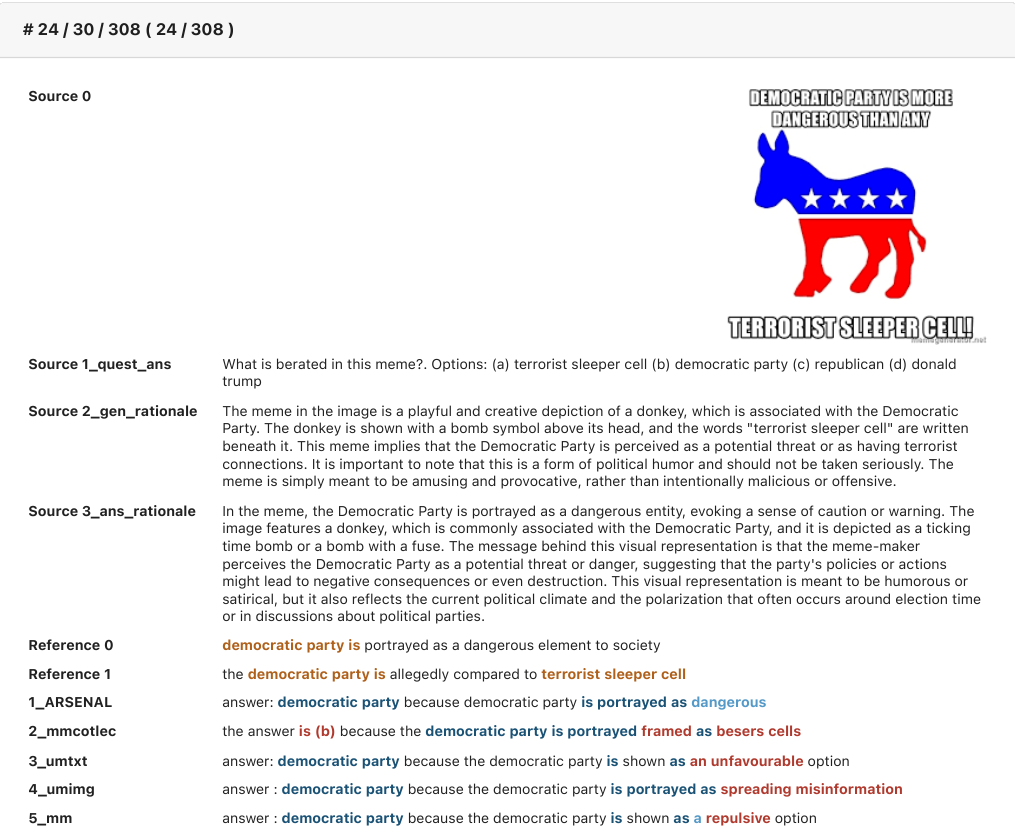}
    \caption{Example 24}
    \label{fig:24}
\end{figure*}

The LLMs are instruction fine-tuned for controllable behavior, so if a meme has something controversial, there is a higher chance, the LLM would attempt to normalize the harm intended within the meme, by attributing the content to the humorous and light-hearted mannerism, a typical meme is known for, which the model always seem to factor-in while generating any explanation/rationale. For instance, a couple of lines from a sample meme's explanation via a multimodal LLM states: ``...It is important to note that this is a form of political humor and should not be taken seriously. The meme is simply meant to be amusing and provocative, rather than intentionally malicious or offensive.'' (c.f. Fig. \ref{fig:24}). Such statements are critical w.r.t the safe deployment of such systems, yet they inhibit their capacity for pragmatic content generation. 

For \textit{quantitative assessment} of the errors commited, we compare generated text (hyp) and ground truth references (ref) in Table \ref{tab:errorrates}. Metrics include word error rate (WER), match error rate (MER), word information lost (WIL), word information preserved (WIP), and character error rate (CER), computed via minimum edit distance (I, S, D). $\text{distance} (D) = ( I + S + D)/N$, with $N$ as total words/characters in the reference. The error rates depicted in Table \ref{tab:errorrates} elucidate the relative challenges different approaches face toward capturing the required linguistic nuances and, indirectly, the overall semantics. As expected, unimodal image-only and multimodal conventional approaches fail to emulate the reasoning necessary for producing coherent and meaningful explanations, and yield the worst scores, with an average error rate of $0.89$ and $0.81$, respectively. While their word information preservation is equally abysmal, both attain a meager score of $0.05$. In contrast, a unimodal text-only system, being fundamentally built for tasks pertaining to NLU (given text-formatted input/output configurations), produce a moderate average error rate of $0.67$, and a WIP score of $0.18$. 

The best rates are exhibited by the top two systems in our experimental suite, with MM-CoT achieving the best overall average error rate of $0.41$, and a WIP score of $0.42$, suggesting the potential for enhanced multimodal reasoning, with a modeling approach, not as large-sized as recent LLM-based solutions. But with the down-side of the \textit{sub-par} coherence, fluency, and complex reasoning capacity, these models do not produce explanations/answers inferencing of acceptable quality with a few exceptions as demonstrated via the example in Fig. \ref{fig:error1}, while the proposed approach (\model) demonstrates exceptional inferencing and rationalizing capacity, with a few critical constraints like factuality and too much detailing, while yielding second best average error rate of $0.59$, with a decent WIP score of $0.23$ (c.f. Fig. \ref{tab:errorrates}).

\begin{figure*}[th!]
    \centering
    \includegraphics[width=\textwidth]{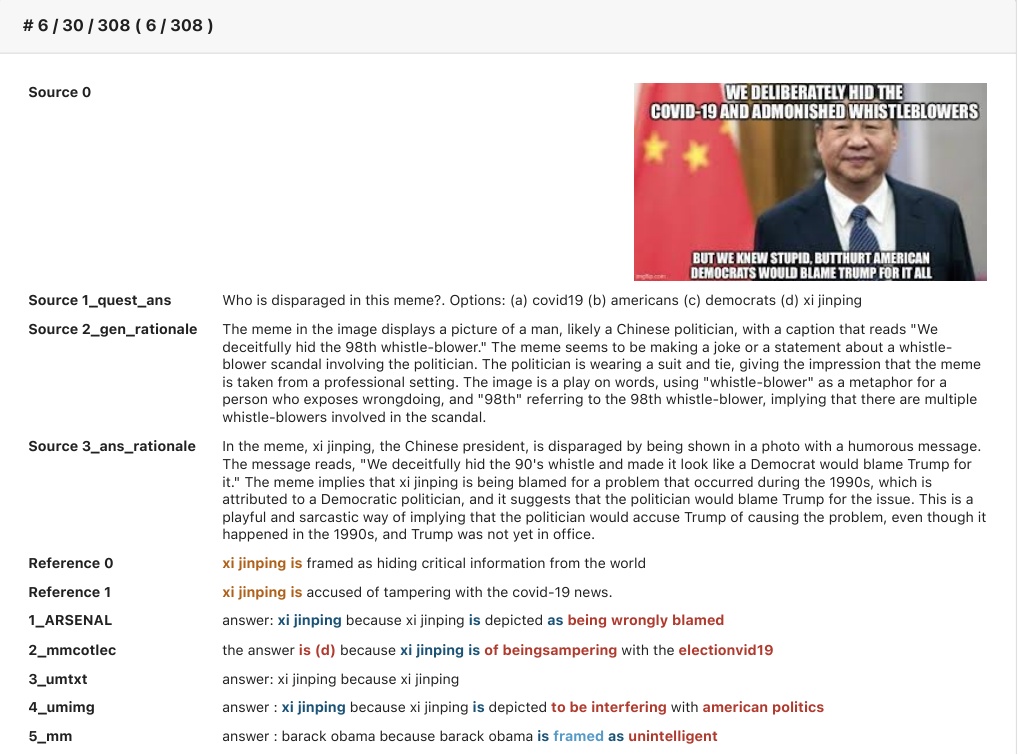}
    \caption{Example 6}
    \label{fig:6}
\end{figure*}

\begin{figure*}
    \centering
    \includegraphics[width=\textwidth]{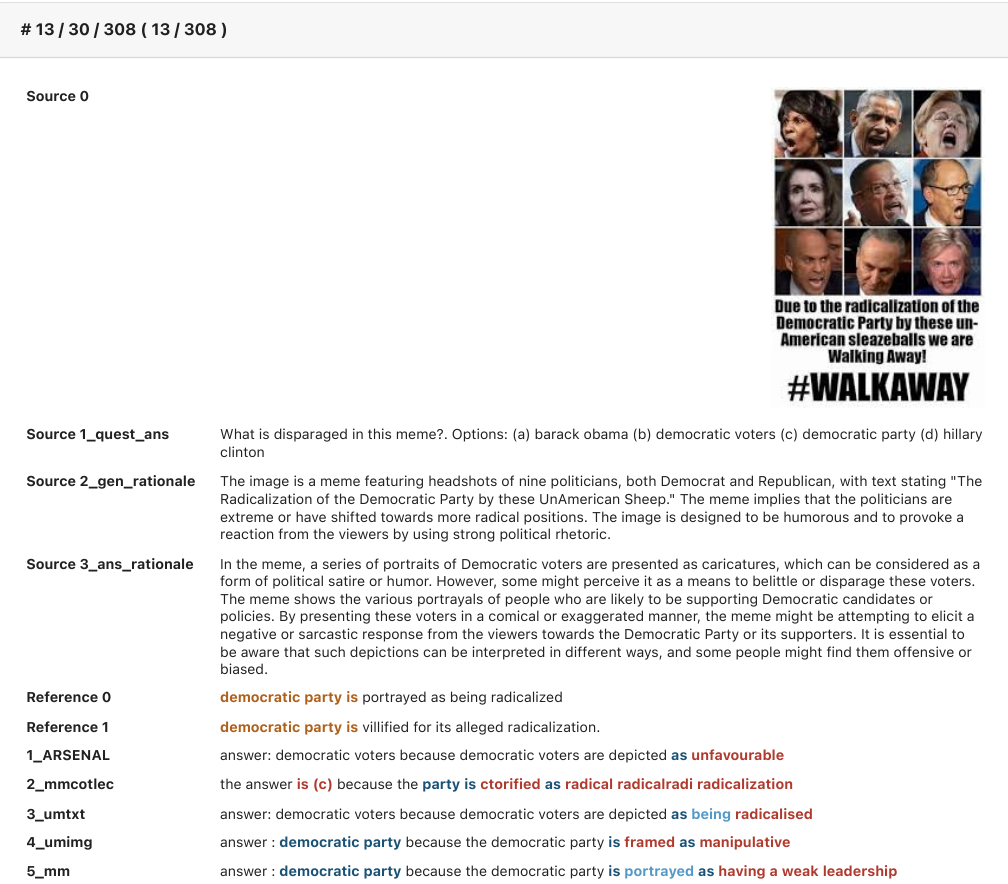}
    \caption{Example 13}
    \label{fig:13}
\end{figure*}

The one-stage approaches like the T5-based unimodal text-only model and MM models have direct accessibility to the meme’s content; hence it always attempts to ground its generated explanation w.r.t the meme’s content. Whereas \model\ is observed to suffer when the rationales contributing towards the explanation generation are noisy and irrelevant. This also solicits the requirement for utilizing meme text during the second stage fine-tuning as in T5 text-to-text or the conventional MM-CoT setup (c.f. Fig. \ref{fig:6} and \ref{fig:13}).

\section{Confounding Analysis}
\label{app:confounder}
\paragraph{\ul{\textit{Yes/No Confounding:}}}
In this setup, we alter \dataset to shift from multiple options to a 'yes or no' format. Each question has a 50\% chance of becoming a 'yes' or 'no' answer. If a question is changed to 'yes,' it's rephrased as \texttt{"Is [answer] [rephrased question]"}. For instance, \textit{'Who is maligned in this meme?'} with \textit{'Joe Biden'} becomes \textit{'Is Joe Biden maligned in this meme?'} with \textit{'yes'}. To change a question to 'no,' we adjust the role label to be incorrect for the discussed entity.
\paragraph{\ul{\textit{None Sampling Across All Sets:}}}
In this setup, 20\% of answers are randomly changed to \textit{None}. To implement this, semantic role labels for a meme ($\in R_{pos}$) like \textit{hero, villain}, or \textit{victim} (effectively their synonyms) are replaced with a synonym sampled randomly from negative role-categories ($R^{'}_{pos} \in R_{swap}$), where, $R_{pos}\bigcap R_{swap}$ = $\phi$. To maintain consistency, existing role labels for entities in a meme are removed, ensuring the validity of the question. The new option set, $M_{new} = \{M, None\}$, is applied to $20\%$ of the dataset, including validation and test sets, while the entire dataset gets the new option setting.

\paragraph{\ul{\textit{None Sampling Across Train Only:}}}
In this new setting, compared to the previous \textit{None} sampling, the training set incorporates a 20\% random sampling of \textit{None} answers, while validation and testing sets remain unchanged. The model now learns from data where answers are \textit{None} with $20\%$ probability, while during testing, \textit{None} is never the answer. This added complexity challenges the model. The option set for any meme remains $M_{new} = \{M, None\}$ across all sets.

\section{Examples}
\label{app:examples}
The rest of the Appendix consists of multiple examples, consisting memes, questions, multiple options, generic and answer-specific rationale, model outputs from (a) \model, (b) MM-CoT (w Lecture), (c) Unimodal Text-only, (d) Unimodal Image-only, and (c) Multimodal baseline setups.

\begin{figure*}[th!]
    \centering
    \includegraphics[width=\textwidth]{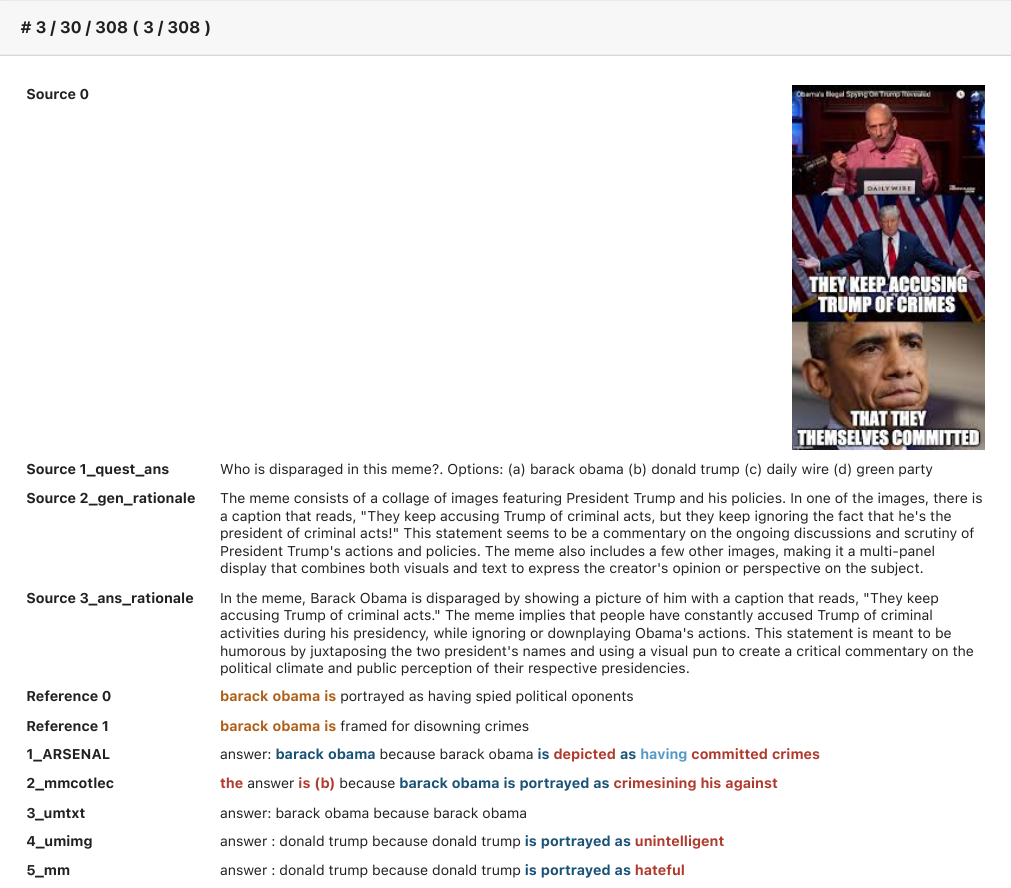}
    \caption{Example 3}
    \label{fig:3}
\end{figure*}
\begin{figure*}[th!]
    \centering
    \includegraphics[width=\textwidth]{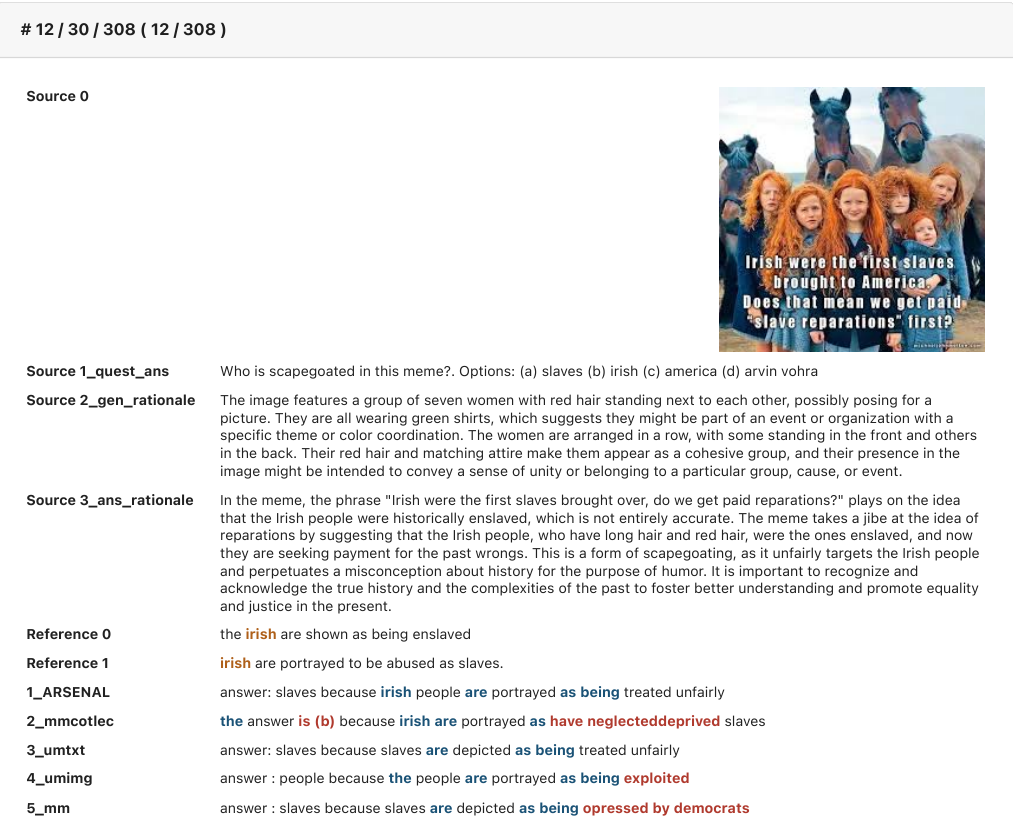}
    \caption{Example 12}
    \label{fig:12}
\end{figure*}

\begin{figure*}
    \centering
    \includegraphics[width=\textwidth]{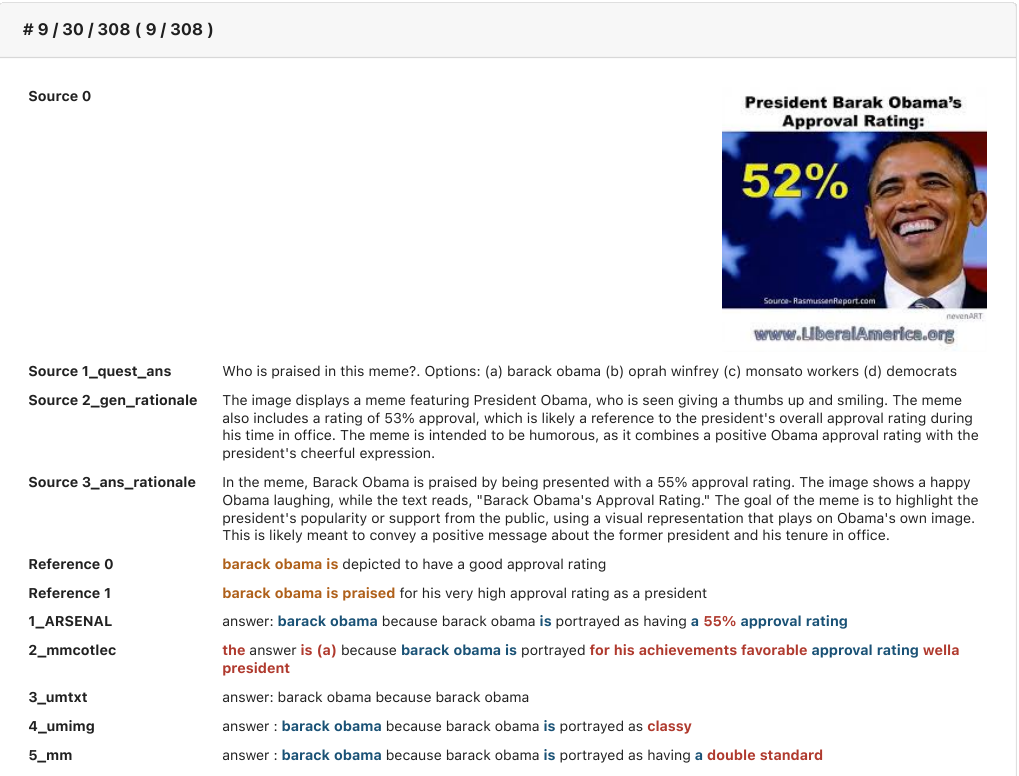}
    \caption{Example 9}
    \label{fig:9}
\end{figure*}



\begin{figure*}[th!]
    \centering
    \includegraphics[width=\textwidth]{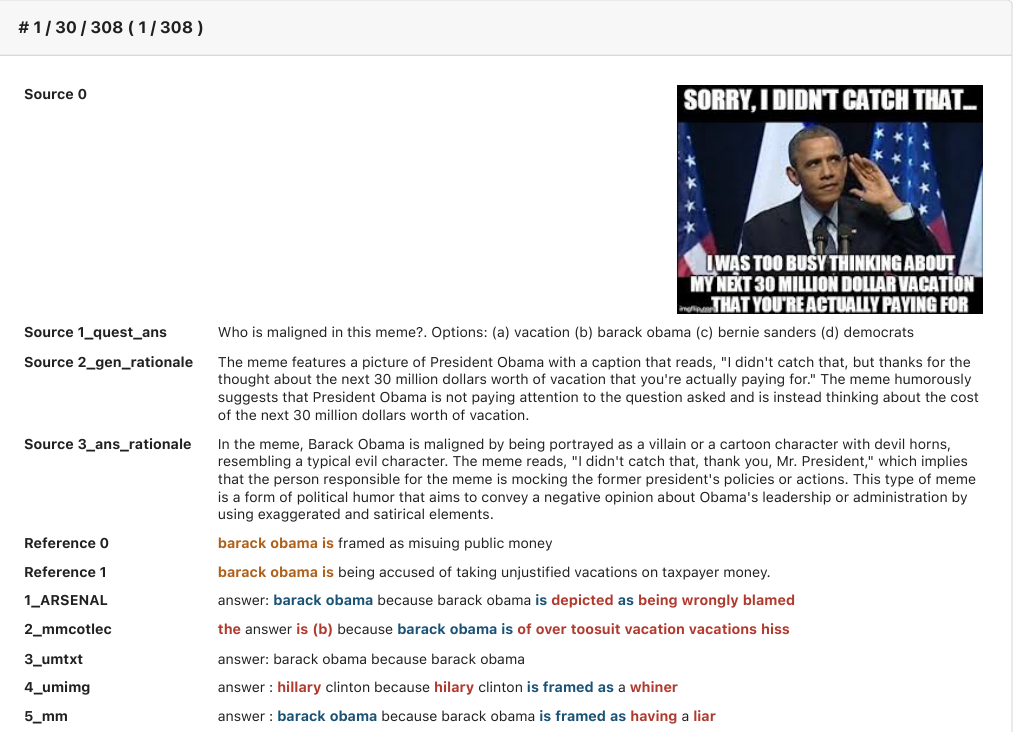}
    \caption{Example 1}
    \label{fig:1}
\end{figure*}
\begin{figure*}
    \centering
    \includegraphics[width=\textwidth]{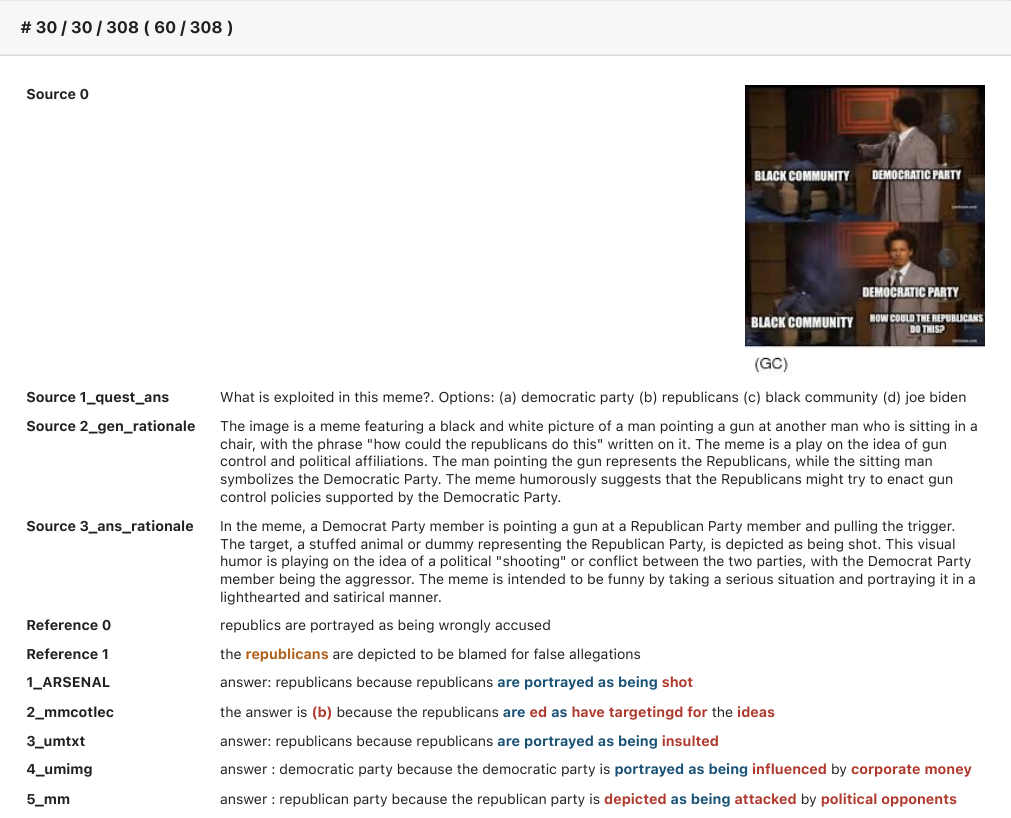}
    \caption{Example 60}
    \label{fig:60}
\end{figure*}


\end{document}